%% file: main.tex
\documentclass[10pt,twocolumn,letterpaper]{article}

\usepackage{ulem}

\usepackage{cvpr}              % To produce the CAMERA-READY version
\definecolor{cvprblue}{rgb}{0.21,0.49,0.74}
\usepackage[pagebackref,breaklinks,colorlinks,allcolors=cvprblue]{hyperref}
\usepackage{algorithm}
\usepackage{algorithmic}
\usepackage{amsmath}
\usepackage{graphicx} % DO NOT CHANGE THIS
\usepackage{subcaption}
\usepackage{booktabs}
\usepackage{multirow}
\usepackage{array}
\usepackage{pifont}
\usepackage{lscape}
\usepackage{tabularx}
\usepackage{amssymb}
\usepackage{pifont}
\usepackage{booktabs}
\usepackage[dvipsnames]{xcolor}
\definecolor{mygray}{RGB}{220, 220, 220}
\usepackage[marginal]{footmisc}

\newcommand{\cmark}{\ding{51}}% Checkmark
\title{MITracker: Multi-View Integration for Visual Object Tracking}

\author{
    Mengjie Xu$^{1*}$ \quad Yitao Zhu$^{1*}$ \quad Haotian Jiang$^{1}$ \quad Jiaming Li$^{1}$ \quad Zhenrong Shen$^{2}$ \\
    Sheng Wang$^{1,2}$ \quad Haolin Huang$^{1}$ \quad Xinyu Wang$^{1}$ \quad Qing Yang$^{1,3}$ \quad Han Zhang$^{1,3}$ \quad Qian Wang$^{1,3 \dag}$ \\ 
    \textsuperscript{1}School of Biomedical Engineering \& State Key Laboratory of\\Advanced Medical Materials and Devices, ShanghaiTech University\\
    \textsuperscript{2}School of Biomedical Engineering, Shanghai Jiao Tong University\\
    \textsuperscript{3}Shanghai Clinical Research and Trial Center\\
    \tt\small \{xumj2023, zhuyt, jianght2023, lijm2024\}@shanghaitech.edu.cn \\ \tt\small \{zhenrongshen, wsheng\}@sjtu.edu.cn \\ \tt\small \{huanghl2023, wangxy42023, yangqing, zhanghan2, qianwang\}@shanghaitech.edu.cn
}

\begin{document}
\maketitle

\begin{abstract}
Multi-view object tracking (MVOT) offers promising solutions to challenges such as occlusion and target loss, which are common in traditional single-view tracking. However, progress has been limited by the lack of comprehensive multi-view datasets and effective cross-view integration methods. To overcome these limitations, we compiled a \textbf{M}ulti-\textbf{V}iew object \textbf{Track}ing (MVTrack) dataset of 234K high-quality annotated frames featuring 27 distinct objects across various scenes. In conjunction with this dataset, we introduce a novel MVOT method, \textbf{M}ulti-View \textbf{I}ntegration \textbf{Tracker} (MITracker), to efficiently integrate multi-view object features and provide stable tracking outcomes. MITracker can track any object in video frames of arbitrary length from arbitrary viewpoints. The key advancements of our method over traditional single-view approaches come from two aspects: (1) MITracker transforms 2D image features into a 3D feature volume and compresses it into a bird’s eye view (BEV) plane, facilitating inter-view information fusion; (2) we propose an attention mechanism that leverages geometric information from fused 3D feature volume to refine the tracking results at each view. MITracker outperforms existing methods on the MVTrack and GMTD datasets, achieving state-of-the-art performance. The code and the new dataset will be available at \href{https://mii-laboratory.github.io/MITracker/}{mii-laboratory.github.io/MITracker}.
\end{abstract}
\footnote{\textsuperscript{*} These authors contributed equally. \textsuperscript{\dag} Corresponding author.\\}
\input{sections/Introduction}
\input{sections/RelatedWork}
\input{sections/Dataset}

\input{sections/Method}
\input{sections/Experiments}

\section{Conclusion}
In this study, we address key challenges such as occlusion and target loss in MVOT by making two significant contributions: (1) MVTrack, a comprehensive dataset with 234K high-quality annotations across diverse scenes and object categories, and (2) MITracker, a novel visual tracking method that effectively integrates multi-view object features. MITracker achieves SOTA results on MVTrack and GMTD datasets, demonstrating its ability to provide stable and reliable tracking across different viewpoints and video durations. Our contributions lay the foundation for future advancements in MVOT, enabling the development of more robust and accurate tracking systems for real-world scenarios.

\section{Acknowledgments}
This work was partially supported by STI 2030-Major Projects (2022ZD0209000) and HPC Platform of ShanghaiTech University.

{
    \small
    \bibliographystyle{ieeenat_fullname}
    \bibliography{main}
}

\input{sections/X_suppl}

\end{document}

%% file: sections/Introduction.tex
\section{Introduction}
\begin{figure}[h]
\centering
\includegraphics[width=1.0\columnwidth]{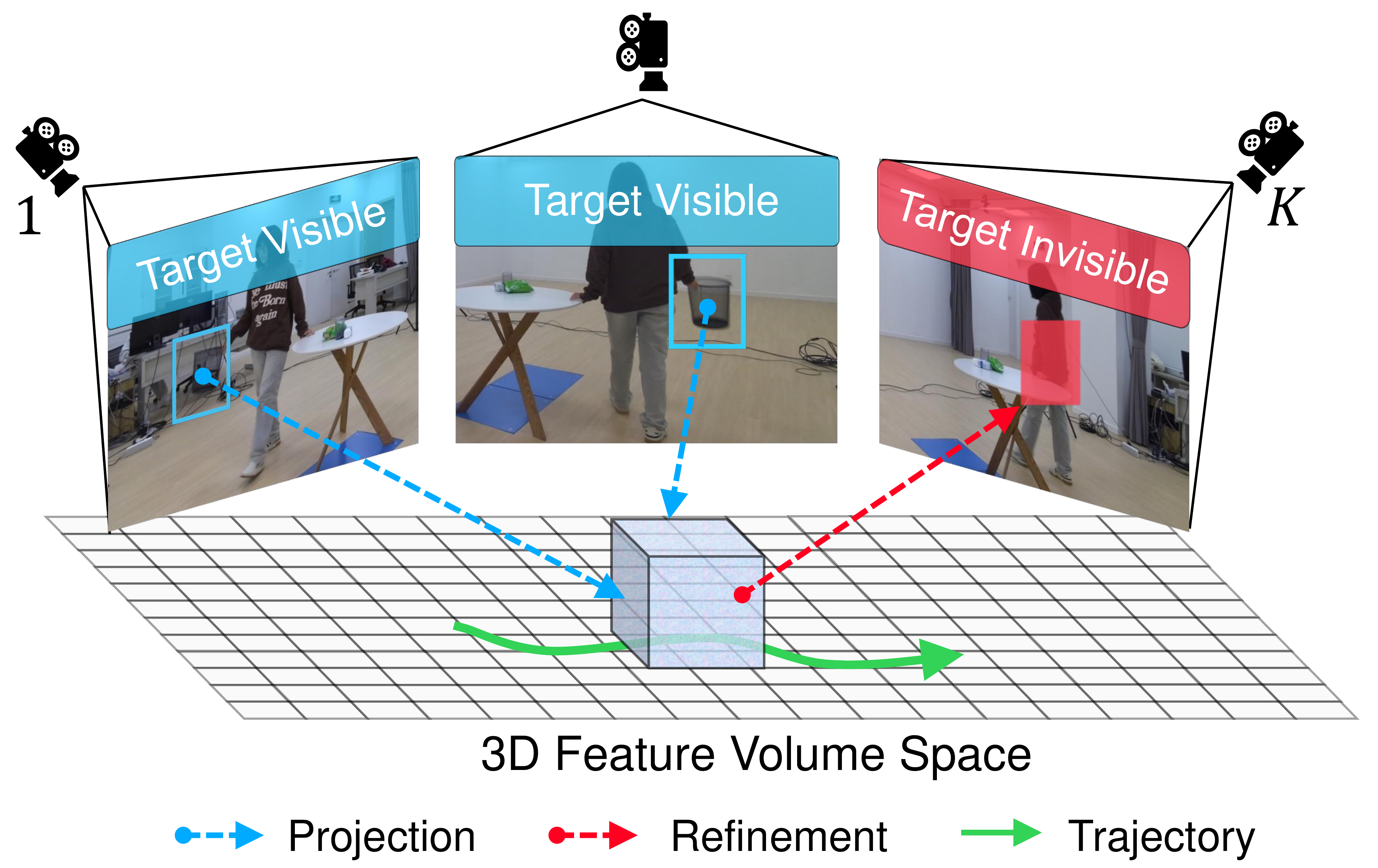} 
\caption{Overview of MITracker's multi-view integration mechanism. Given $K$ camera views, our method projects features from views with visible targets into a 3D feature volume space, which is then used to refine tracking in views where the target is occluded.}
\label{fig: intro_fig}
\end{figure}

Visual object tracking, a core computer vision task, involves estimating class-agnostic target positions across video sequences. 
This technique is crucial for applications such as augmented reality and autonomous driving, where it is essential to continuously monitor and predict the trajectories of various objects within dynamic environments. 
Despite notable advances in single-view tracking through Siamese networks \cite{li2018high, danelljan2020probabilistic} and transformers \cite{cui2022mixformer, cai2023robust, zheng2024odtrack}, significant challenges persist – particularly occlusions, appearance changes, and target loss. 
While approaches like RTracker \cite{huang2024rtracker} attempt to address these challenges by determining target loss and detection mechanisms, the inherent limitations of single viewpoint information remain a fundamental constraint.

Multi-camera systems offer a promising solution by leveraging complementary viewpoints to maintain continuous tracking, particularly for handling occlusions through camera overlap \cite{zhu2024muc}. 
However, the development of effective multi-view object tracking (MVOT) faces several critical challenges. 
First, existing multi-view datasets are largely restricted to specific object categories like humans or birds \cite{han2023mmptrack, xiao2023multi}, limiting their applicability for generic object tracking. 
Second, current MVOT approaches \cite{xu2016multi, hou2020multiview, harley2023simple} primarily focus on tracking specific categories of objects using detection and re-identification methods, which are not suitable for class-agnostic object tracking. 
Even when attempting to track generic objects across multiple views, 
researchers have to rely on single-view datasets for training due to the absence of comprehensive multi-view data \cite{wu2020visual}. 
This limitation severely restricts models' ability to understand complex spatial relationships and appearance variations across different viewpoints.

To address these challenges, we first construct a \textbf{M}ulti-\textbf{V}iew object \textbf{Track}ing (MVTrack) dataset.
MVTrack dataset contains 234K frames captured from 3-4 cameras, with precise bounding box (BBox) annotations covering 27 distinct objects across 9 challenging tracking attributes such as occlusion and deformation.
Unlike existing datasets such as GMTD \cite{wu2020visual} which only provides testing data, MVTrack dataset offers both training and evaluation sets, enabling development and validation of MVOT models.

To effectively utilize MVTrack dataset, we propose a novel MVOT method named \textbf{M}ulti-View \textbf{I}ntegration \textbf{Tracker} (MITracker) for tracking any object in video frames of arbitrary length from arbitrary viewpoints.
As illustrated in Figure \ref{fig: intro_fig}, MITracker can integrate multi-view features into a unified 3D feature volume and further refine tracking in occluded views, thus producing robust tracking outcomes.
The framework of MITracker consists of two important modules: \textbf{View-Specific Feature Extraction} and \textbf{Multi-View Integration}.
The first module employs a Vision Transformer (ViT) \cite{dosovitskiy2020image} to extract view-specific features of the target object from the current search frame in a streaming manner, where the target object is indicated by a reference frame.
The second module constructs a 3D feature volume by fusing 2D features from multiple views and leveraging bird’s eye View (BEV) guidance, which significantly enhances the model's spatial understanding. 
This 3D feature volume is then deployed in spatial-enhanced attention to improve tracking accuracy.
MITracker allows for the maintenance of stable tracking results and demonstrates strong recovery capabilities in challenging cases such as occlusions and out of view objects.

In summary, our main contributions are as follows:

\begin{itemize}
    \item We introduce MVTrack, a large-scale multi-view tracking dataset containing 234K frames from 3-4 calibrated cameras. It has precise BBox annotations of 27 object categories across 9 challenging tracking attributes, which provides the first comprehensive benchmark for training class-agnostic MVOT methods and enriches the approaches for evaluating these methods. 
    \item We propose MITracker, a novel multi-view tracking method that constructs BEV-guided 3D feature volumes to enhance spatial understanding and utilize a spatial-enhanced attention mechanism to enable robust recovery from target loss in specific views.
    \item Our extensive experiments demonstrate that MITracker achieves state-of-the-art (SOTA) performance on both MVTrack and GMTD datasets, improving recovery rate from 56.7\% to 79.2\% to reduce target loss in challenging scenarios.
\end{itemize}

%% file: sections/RelatedWork.tex
\section{Related Work}

\input{tables/dataset/benchmark}
\subsection{Visual Object Tracking}
Visual object tracking has garnered significant research interest, leading to many breakthroughs. Numerous single-view datasets \cite{OTB2015, kiani2017need, kristan2016novel, muller2018trackingnet, fan2019lasot, huang2019got, wang2021towards, hu2022global, peng2024vasttrack} span a wide range of categories, aimed at enhancing models' ability to track arbitrary objects. With the expansion of these datasets, single-view tracking methods have also advanced rapidly. 
Early approaches based on Siamese networks \cite{li2018high, danelljan2020probabilistic} use CNNs to extract features from reference and search regions, establishing a linear relationship between them. 
More recent works have incorporated transformers for enhanced feature extraction \cite{cui2022mixformer, ye2022joint}, while others introduce attention modules to enable nonlinear relationships \cite{chen2021transformer}. 
However, these methods lack temporal continuity as they process each frame independently.
Algorithms like dynamic template updating \cite{chen2023seqtrack} and spatio-temporal trajectory tracking \cite{wei2023autoregressive, zheng2024odtrack} have shown promising results in addressing this issue.
Despite these advancements, recovering from target loss remains a significant challenge.

To re-track the target after a tracking failure, RTracker \cite{huang2024rtracker} leverages a tree-structured memory system to detect target loss and a dedicated detector for self-recovery. However, this approach is constrained by its complex design and the detector's reliance on specific categories.
Single-view tracking suffers from inherent limitations due to its restricted field of view, which is an inevitable challenge. In contrast, GMT \cite{wu2020visual} incorporates multi-view tracking within a single-view training framework. This limits its capacity to effectively model the intricate relationships between multi-view appearances and background contexts in the real world.

\subsection{Multi-View Object Tracking}
MVOT provides more comprehensive information about the target, effectively addressing issues such as occlusion. 
To leverage multi-view information, various fusion strategies have been developed for target association across viewpoints.
Some approaches establish multi-view relationships by projecting detection results onto a BEV plane \cite{xu2016multi}. 
However, this method is prone to detection errors, especially with occlusion.
% To tackle this issue, MVDet \cite{hou2020multiview} improves it by incorporating multi-view information at the detection stage (known as early fusion) through feature projection onto the ground plane, enabling the model to capture richer interaction information across views. 
To tackle this issue, methods \cite{hou2020multiview, hou2021multiview} improve it by incorporating multi-view information at the detection stage (known as early fusion) through feature projection onto the ground plane, enabling the model to capture richer interaction information across views. 
% Building on this approach, Simple-BEV \cite{harley2023simple} maps features into 3D space at multiple heights, which reduces distortions caused by ground plane projection.
Building on these approaches, methods such as \cite{song2021stacked, harley2023simple, teepe2024earlybird} map features into 3D space at multiple heights, which reduces distortions caused by ground plane projection.

While these methods enhance multi-view tracking, existing multi-view datasets are relatively scarce and often limited to specific target categories, such as pedestrians \cite{chavdarova2018wildtrack, han2023mmptrack, hao2024divotrack}. 
This results in a reliance on detection outcomes, limiting the ability to track arbitrary objects. 
GMTD \cite{wu2020visual} expands the target to multiple categories, but its scale remains small, primarily designed for evaluation purposes. There is a pressing need for a multi-view tracking dataset capable of handling arbitrary objects.

%% file: tables/dataset/benchmark.tex
\begin{table*}[t]
\centering
\resizebox{\textwidth}{!}{
\begin{tabular}{l c c c c c c c c c c c c}
\hline
\multirow{2}{*}{\textbf{Benchmark}} & \multirow{2}{*}{\textbf{Aim}}& \multirow{2}{*}{\textbf{Camera}} & \multirow{2}{*}{\textbf{Class}} & \textbf{Total} & \multirow{2}{*}{\textbf{Videos}} & \textbf{Mean}  & \textbf{Absent} & \multirow{2}{*}{\textbf{Att.}}  &  \multirow{2}{*}{\textbf{Overlap}} & \multirow{2}{*}{\textbf{Move}} & \multirow{2}{*}{\textbf{Calib.}} \\
  & & & & \textbf{Frames} & &  \textbf{Frames}&\textbf{label} & &  & & &\\
\hline
OTB 2015 \cite{OTB2015} &Eva.& 1 & 16 & 59K& 100 & 590 & \ding{55} &11  &-&-&- \\
NfS \cite{kiani2017need} & Eva.& 1 & 17 & 383K& 100 & 3,830 &\ding{55} & 9  &-&-&-\\
VOT 2017 \cite{kristan2016novel} & Eva.& 1&24 & 21K& 60 & 356  & \ding{55} & 24 &-&-&- \\
TrackingNet \cite{muller2018trackingnet} & Tra./Eva.&1 & 27 &14.43M& 30,643 & 471  & \ding{55} & 15 &-&-&-\\
LaSOT \cite{fan2019lasot} &Tra./Eva.&1 &  70 & 3.52M& 1,400 & 2,053  & \ding{51} & 14 &-&-&-\\
GOT-10k \cite{huang2019got} & Tra./Eva.& 1 & 563 &1.45M& 9,935 & 149  & \ding{51} & 6 &-&-&-\\
TNL2K \cite{wang2021towards} & Tra./Eva.& 1 & - & 1.24M& 2,000 & 622  & \ding{51} & 17 &-&-&-\\
VideoCube \cite{hu2022global} & Tra./Eva.&  1& 89 &7.46M& 500 & 4,008  & \ding{51} & 12 &-&-&-\\
VastTrack \cite{peng2024vasttrack} & Tra./Eva.& 1 & 2,115 & 4.20M& 50,610 & 83  & \ding{51} & 10 &-&-&-\\
\hline
CAMPUS$^\dagger$ \cite{xu2016multi} & Eva.& 4 & 1 & 83K& 16& 5,188& - & - & \ding{51} & \ding{55} & \ding{55} \\
Wildtrack$^\dagger$ \cite{chavdarova2018wildtrack} &  Eva.& 7 & 1 & 2.80K&  7 &  401 & - & - & \ding{51} & \ding{55} & \ding{51}\\
MMPTRACK$^\dagger$ \cite{han2023mmptrack} & Tra./Eva.& 4-6& 1 & 2.98M& -&  -& -& -& \ding{51} & \ding{55} & \ding{51}\\
DIVOTrack$^\dagger$ \cite{hao2024divotrack} & Tra./Eva.& 3 & 1 & 81K&75& 1,080& -& -&\ding{51}& \ding{51}& \ding{55}\\
% \hline
GMTD \cite{wu2020visual} & Eva.& 2-3 & 8 & 18K&  23 & 764 & \ding{55}& 6 & \ding{51}& \ding{51}& \ding{55}\\
\textbf{MVTrack (Ours)} & Tra./Eva.& 3-4 & 27 & 234K& 260 & 901  & \ding{51} & 9 & \ding{51}&\ding{55} &\ding{51} \\
\hline
\end{tabular}
}
\caption{Comparison of current datasets for object tracking. The upper part of the table focuses on single-view datasets, while the lower part is dedicated to multi-view datasets. Datasets marked with $^\dagger$ are designed for multi-object tracking, the others are for visual object tracking. `Tra.' and `Eva.' indicate training and evaluation, respectively. `-' denotes not available, `Att.' stands for attributes, `Overlap' refers to the overlapping of multi-view images, `Move' indicates the movement of the camera position, and `Calib.' represents calibration. }
\label{tab: benchmark dataset comparison}
\end{table*}

%% file: sections/Dataset.tex
\begin{figure}[t]
    \centering
    \begin{subfigure}[b]{1.0\columnwidth}
        \centering
        \includegraphics[width=\columnwidth]{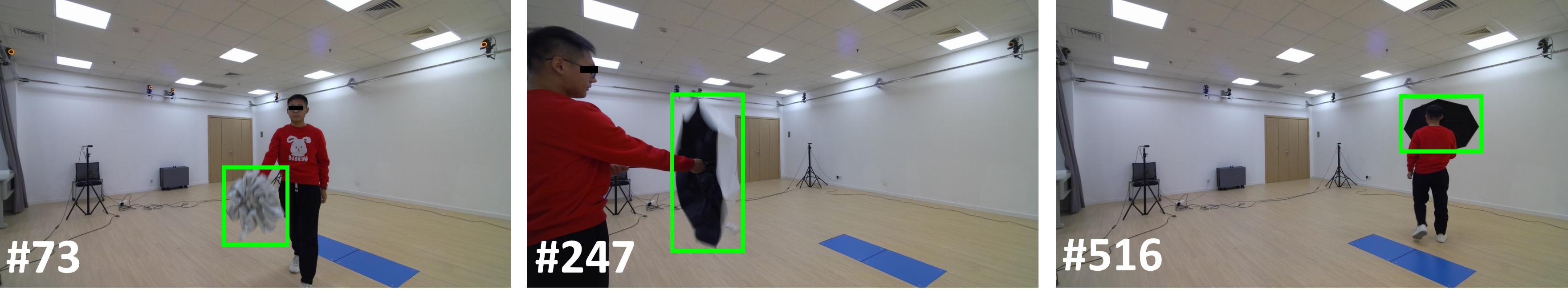}
        \caption{\textit{umbrella1-1}: Deformation, Aspect Ratio Change and Scale Variation.}
        \label{fig: data-umbrella}
    \end{subfigure}
    
    \begin{subfigure}[b]{1.0\columnwidth}
        \centering
        \includegraphics[width=\columnwidth]{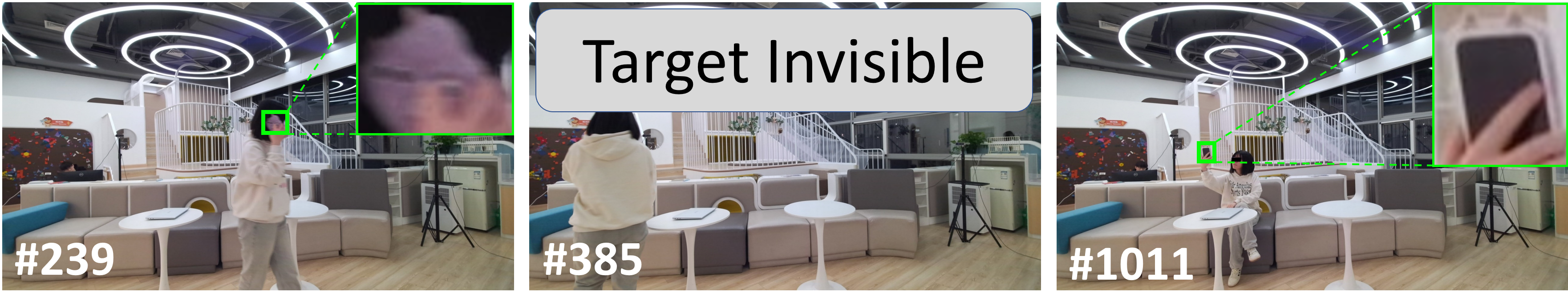}
        \caption{\textit{phone3-3}: Low Resolution, Fully Occlusion and Partial Occlusion.}
        \label{fig: data-phone}
    \end{subfigure}

    \begin{subfigure}[b]{1.0\columnwidth}
        \centering
        \includegraphics[width=\columnwidth]{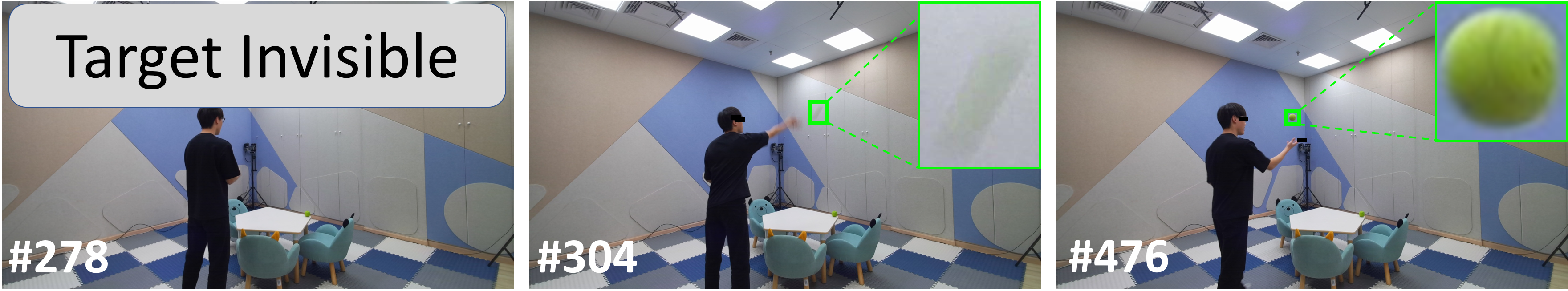}
        \caption{\textit{tenis5-1}: Out of View, Motion Blur and Background Clutter.}
        \label{fig: data-tennis}
    \end{subfigure}
    
    \caption{Example sequences, annotations, and their corresponding tracking attributes in the MVTrack dataset.}
    \label{fig:data-vis}
\end{figure}

\section{MVTrack Dataset}
MVTrack dataset is designed to fill the gaps in the field of MVOT and has received approval for data collection from an Institutional Review Board.
As shown in Table \ref{tab: benchmark dataset comparison}, compared to single-view datasets, we maintain competitive class diversity while adding multi-view capabilities. Compared to MVOT datasets, we provide significantly richer object categories (27 vs 1-8 classes) and more videos (260) with practical camera setups (3-4 views). MVTrack dataset is the only dataset that combines multi-view tracking, rich object categories, absent label annotations, and calibration information.

\textbf{Data Collection.} 
We employ a multi-camera system for data collection, consisting of 3 or 4 time-synchronized Azure Kinect cameras. 
All video sequences are recorded at a resolution of 1920$\times$1080 with 30 FPS.
These cameras are positioned to ensure multiple overlapping views, and their intrinsic parameters are provided by the manufacturer. 
The extrinsic parameters are obtained through calibration and finely adjusted using MeshLab \cite{meshlab, LocalChapterEvents:ItalChap:ItalianChapConf2008:129-136}. 
With this calibration information, we set the central point of the scene as the origin of the world coordinate system, aligning all viewpoints to this unified coordinate system.

\textbf{Data Annotation.}
MVTrack dataset provides frame-level annotations, including 2D object BBoxes and ground coordinate annotations in a unified coordinate system (i.e., BEV annotations).
Following an annotation strategy similar to LaSOT \cite{fan2019lasot}, where for each visible frame, an axis-aligned BBox tightly encloses the target, and an `invisible' label is assigned for the invisible target. 
The BBox annotations are generated semi-automatically, with trackers \cite{xie2024autoregressive, chen2023seqtrack, zheng2024odtrack} used for initial labeling. 
The machine-generated annotations are then manually adjusted and double-checked for accuracy.
Subsequently, using camera calibration parameters, the 2D object BBoxes from multiple viewpoints are projected into the unified coordinate system to compute the BEV coordinates.

\begin{figure*}[t]
\centering
\includegraphics[width=1\textwidth]{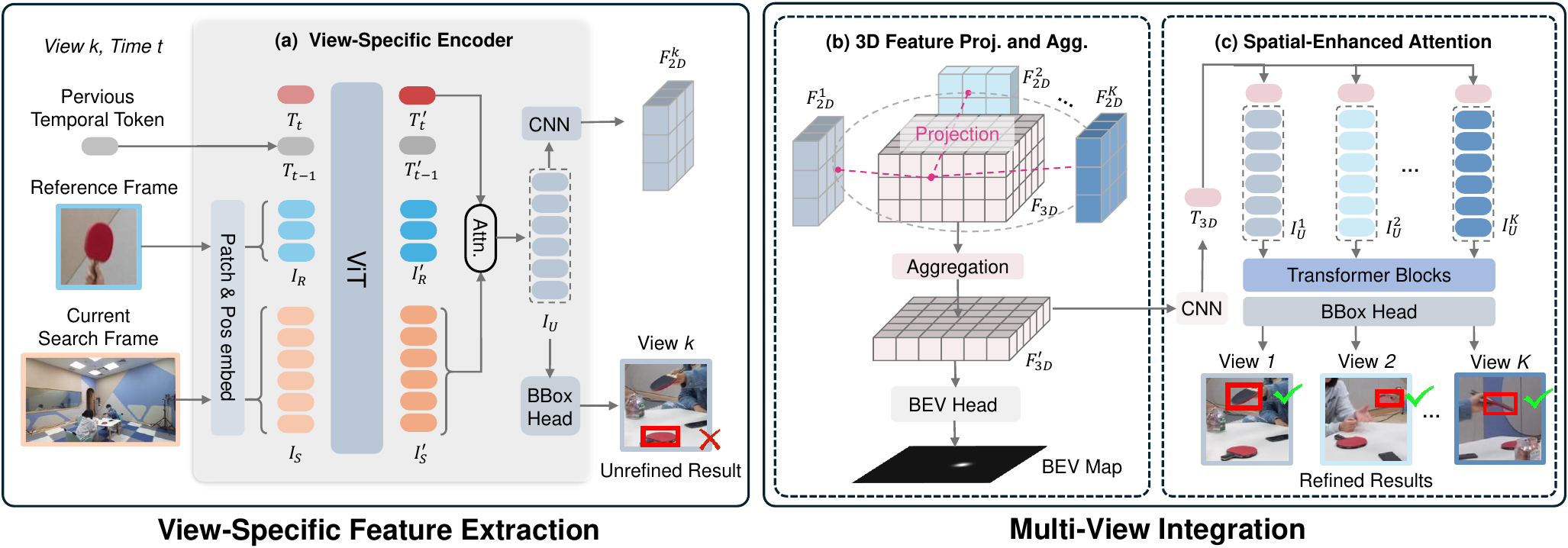} 
\caption{The framework of MITracker. (a) The view-specific feature extraction module employs a ViT that utilizes temporal tokens to process each view independently, outputting unrefined results that can be further improved by multi-view information. The multi-view integration module contains (b) 3D feature volume construction that aggregates features into 3D space with BEV guidance and (c) spatial-enhanced attention that refines tracking results by 3D spatial information.}
\label{fig: method_pipeline}
\end{figure*}

\textbf{Challenging Attributes.} 
In our dataset, we particularly focus on 9 common tracking challenges to better assess tracker performance: Background Clutter, Motion Blur, Partial Occlusion, Full Occlusion, Out of View, Deformation, Low Resolution, Aspect Ratio Change, and Scale Variation. 

More specifically, Figure \ref{fig:data-vis} illustrates three challenging samples in the MVTrack dataset. Figure \ref{fig: data-umbrella} shows significant deformation and scale changes of an umbrella being opened. Figure \ref{fig: data-phone} demonstrates the tracking of small, low-resolution objects like a mobile phone under full and partial occlusions. Figure \ref{fig: data-tennis} highlights the impact of fast motion causing blur when tracking a tennis ball. These attributes can significantly aid in training the model to achieve more robust results.

\textbf{Statistical Analysis.} 
MVTrack dataset consists of five indoor scenes, captured with a total of ten sets of calibration parameters. 
It covers 27 everyday objects, ranging from small objects like pens to larger objects such as umbrellas. 
The dataset includes 68 sets of multi-view data, comprising 260 videos and a total of 234,430 frames. 

We divide the dataset into training, validation, and testing sets. The training set consists of 196 videos and 180K frames, while the validation set contains 30 videos and 28K frames. The testing set comprises 34 videos and 26K frames. We include an unseen scene in the validation and testing sets that are distinct from the scenes in the training set. Furthermore, the testing set includes both object categories that appear in the training set and new object categories not present during training. This enables evaluation of the model's performance across various targets and settings.

More details about MVTrack dataset are provided in the Appendix.

%% file: sections/Method.tex
\section{MITracker}
We propose MITracker, a novel multi-view tracking framework that robustly tracks class-agnostic objects across multiple camera views. As illustrated in Figure \ref{fig: method_pipeline}, MITracker consists of two main components: (1) a \textbf{view-specific feature extraction} module (Sec. \ref{subsec: View-specific Feature Extraction}) that encodes frame features and generates single-view tracking results in a streaming fashion, and (2) a \textbf{multi-view integration} module (Sec. \ref{subsec: Multi-view Integration}) that fuses multi-view features with BEV guidance and refines view-specific feature with a spatial-enhanced attention mechanism.

\subsection{View-Specific Feature Extraction}
\label{subsec: View-specific Feature Extraction}
As shown in Figure \ref{fig: method_pipeline}a, this module processes the video stream from a specific viewpoint \(k\), and extracts target-aware features in the search frame at a timepoint \(t\) based on the reference frame that indicates the target object.

\textbf{View-Specific Encoder.}
We employ ViT as the backbone of our view-specific encoder.
The visual inputs of the view-specific encoder consist of a search frame $S \in \mathbb{R}^{3 \times H_s \times W_s}$ and a reference frame $R \in \mathbb{R}^{3 \times H_r \times W_r}$. 
As the transformer block processes a series of tokens, we segment the frames into non-overlapping patches with $p \times p$ resolution.
The search and reference frames are individually embedded into a token sequence, represented by $I_S \in \mathbb{R}^{N_s \times D}$ and $I_R \in \mathbb{R}^{N_r \times D}$, where $D$ is the hidden dimension, $N_s = \frac{H_s W_s}{p^2}$ is the number of search tokens, and $N_r = \frac{H_r W_r}{p^2}$ is the number of reference tokens.

To ensure temporal continuity between frames, akin to the method utilized in ODTrack \cite{zheng2024odtrack}, two specialized temporal tokens are also included in the inputs of the view-specific encoder to facilitate the propagation of temporal information. 
Specifically, at any given time \( t \), a learnable token \( T_t \) is randomly initialized, which is designed to capture temporal information of the current frame.
Concurrently, we incorporate a token \( T_{t-1} \) that carries temporal information from the preceding frame, which leverages historical features to enhance tracking accuracy and continuity.
The input token sequence of our view-specific encoder can be formulated as the composition of the visual and temporal tokens $f = [T_{t}, T_{t-1}, I_{R}, I_{S}]$, while the output token sequence is denoted as $f' = [T'_{t}, T'_{t-1}, I'_{R}, I'_{S}]$.

After obtaining $f'$, \( T'_{t} \) is used to compute attention weights in conjunction with \( I'_{S} \) to utilize temporal information for adjustments, which can be described as follows:
\begin{equation}
    I_{U} = I^{\prime}_{S}\cdot(I^{\prime}_{S}\times (T^{\prime}_{t})^\top),
\end{equation}
where \( I_U \) represents the extracted feature that encapsulates attention focused on the target object in the search frame.

\textbf{Single-View Tracking Result.}
We employ a BBox head based on the CenterNet architecture \cite{zhou2019objects} to output tracking results from the extracted feature \( I_U \). 
This head comprises three distinct sub-networks, each designed to compute the classification score map, BBox dimensions, and offset sizes, respectively. The highest-scoring position on the classification score map is identified as the target location. This configuration establishes a robust framework capable of effectively handling single-view visual object tracking tasks.

To facilitate further multi-view integration, we also apply convolutional layers to map \( I_U \) to a 2D feature map with original image size, denoted as \( F_{2D} \in \mathbb{R}^{32 \times H_s \times W_s} \).
This establishes a pixel-wise correspondence between the extracted feature and the search image, which is crucial for reconstructing the 3D feature space in the following section.

\subsection{Multi-View Integration}
\label{subsec: Multi-view Integration}
To effectively integrate 2D feature maps $F_{2D}^{1}, F_{2D}^{2}, ..., F_{2D}^{K}$ from $K$ viewpoints, we project them into a 3D feature space and then aggregate them under the supervision of BEV guidance. 
Finally, we embed the aggregated feature to a 3D-aware token to refine all view-specific features $I_{U}^{1}, I_{U}^{2}, ..., I_{U}^{K}$ via spatial-enhanced attention, thus producing stable tracking results across different viewpoints.

\textbf{3D Feature Projection.}
As illustrated in Figure \ref{fig: method_pipeline}b,  we construct a 3D feature volume of size $X \times Y \times Z$, where $(X, Y)$ represents the horizontal plane and $Z$ axis denotes the vertical direction following \cite{zhang2022voxeltrack, harley2023simple}. 
For a viewpoint $k$, we project the $(u, v)$ coordinates in $F_{2D}^{k}$ to $(x, y, z)$ coordinates in the 3D feature volume by the formula below:
\begin{equation}
\begin{pmatrix} u \\ v \\ 1 \end{pmatrix} = C_K [ C_R | C_t] \begin{pmatrix} x \\ y \\ z \\ 1 \end{pmatrix},
\label{eq:projection}
\end{equation}
where $C_K$ represents the camera’s intrinsic matrix, $C_R$ denotes the rotation matrix describing the camera’s orientation, and $C_t$ is the translation vector specifying the camera’s position in space. 
Upon establishing the mapping matrix, we implement bilinear sampling to populate the 3D feature volume. 
In scenarios that involve multiple viewpoints, we compute the average of the mapped values from each view to ensure consistency. Consequently, we derive a 3D feature volume represented as \( F_{3D} \in \mathbb{R}^{32 \times X \times Y \times Z} \).

\textbf{3D Feature Aggregation.}
To better integrate multi-view spatial information, we apply 1D convolutional layers to aggregate features along the Z-axis of $F_{3D}$, resulting in \( F'_{3D} \in \mathbb{R}^{32 \times X \times Y} \), thereby consolidating spatial information within the \( (X, Y) \) plane.
Subsequently, a classification head (i.e., BEV head) is employed to generate a BEV score map from $F'_{3D}$.
This BEV map delineates the object positions on the horizontal plane, thereby imposing supervision constraints on information fusion across multiple viewpoints. 
This integrative approach allows for precise localization and mapping within multi-view scenarios.

\textbf{Spatial-Enhanced Attention.}
BEV guidance for the aggregated 3D feature $F'_{3D}$ only implicitly constrains the original single-view output, but it is insufficient to address the potential target loss issue due to the lack of direct supervision on tracking results.
To remedy this, we introduce spatial-enhanced attention to explicitly incorporate $F'_{3D}$ into the tracking process as shown in Figure \ref{fig: method_pipeline}c. 

We first use convolutional layers to embed $F'_{3D}$ into a 3D-aware token $T_{3D} \in \mathbb{R}^{1 \times D}$, which inherits multi-view spatial information. 
For all the $K$ viewpoints, we then individually concatenate $T_{3D}$ with their unrefined features $I_{U}^{1},I_{U}^{2},...,I_{U}^{K}$ produced by the view-specific encoder.
For a viewpoint $k$, a series of transformer blocks take in its composite token sequence $(T_{3D}, I_{U}^{k})$ and refine them using attention mechanisms that leverage fused 3D spatial information.
A final BBox head outputs the refined tracking results, where potential errors such as target loss are corrected.

%% file: sections/Experiments.tex
\section{Experiments}

\input{tables/experiments/compairson}

\subsection{Dataset}
In addition to MVTrack dataset, we use two external datasets for training and evaluation, which are detailed as follows.

\begin{itemize}
    \item \textbf{GOT10K.} GOT-10K \cite{huang2019got} is a large and diverse dataset with a wide range of object categories. Its training set contains 9,335 videos across 480 moving object categories.
    \item \textbf{GMTD.} GMTD \cite{wu2020visual} is a multi-view tracking test set with 10 scenes, captured by 2-3 uncalibrated cameras in indoor and outdoor settings. It includes 6 target types and various tracking challenges.
\end{itemize}

\subsection{Implementation Details}
\textbf{Loss Function.}
For the BBox head, we employ a weighted focal loss \cite{lin2017focal} \(L_{\text{cls}}\)  for classification, along with the generalized intersection over union loss \cite{rezatofighi2019generalized} \(L_{\text{giou}}\) and \(L_1\) loss for BBox regression. Additionally, a focal loss \(L_{\text{bev}}\) is utilized for BEV map supervision.
The overall loss function of the model is formulated as follows:
\begin{equation}
L_{\text{track}} = L_{\text{cls}} + \lambda_{\text{giou}} L_{\text{giou}} + \lambda_{L_1} L_1 + \lambda_{\text{bev}} L_{\text{bev}},
\end{equation}
where \(\lambda_{\text{giou}} = 5\), \(\lambda_{L_1} = 2\), and \(\lambda_{\text{bev}} = 0.1\) are the coefficients that balance the contributions from each loss .

\begin{figure*}[htbp]
	\centering
	\begin{subfigure}{0.66\columnwidth}
		\centering
		\includegraphics[width=\columnwidth]{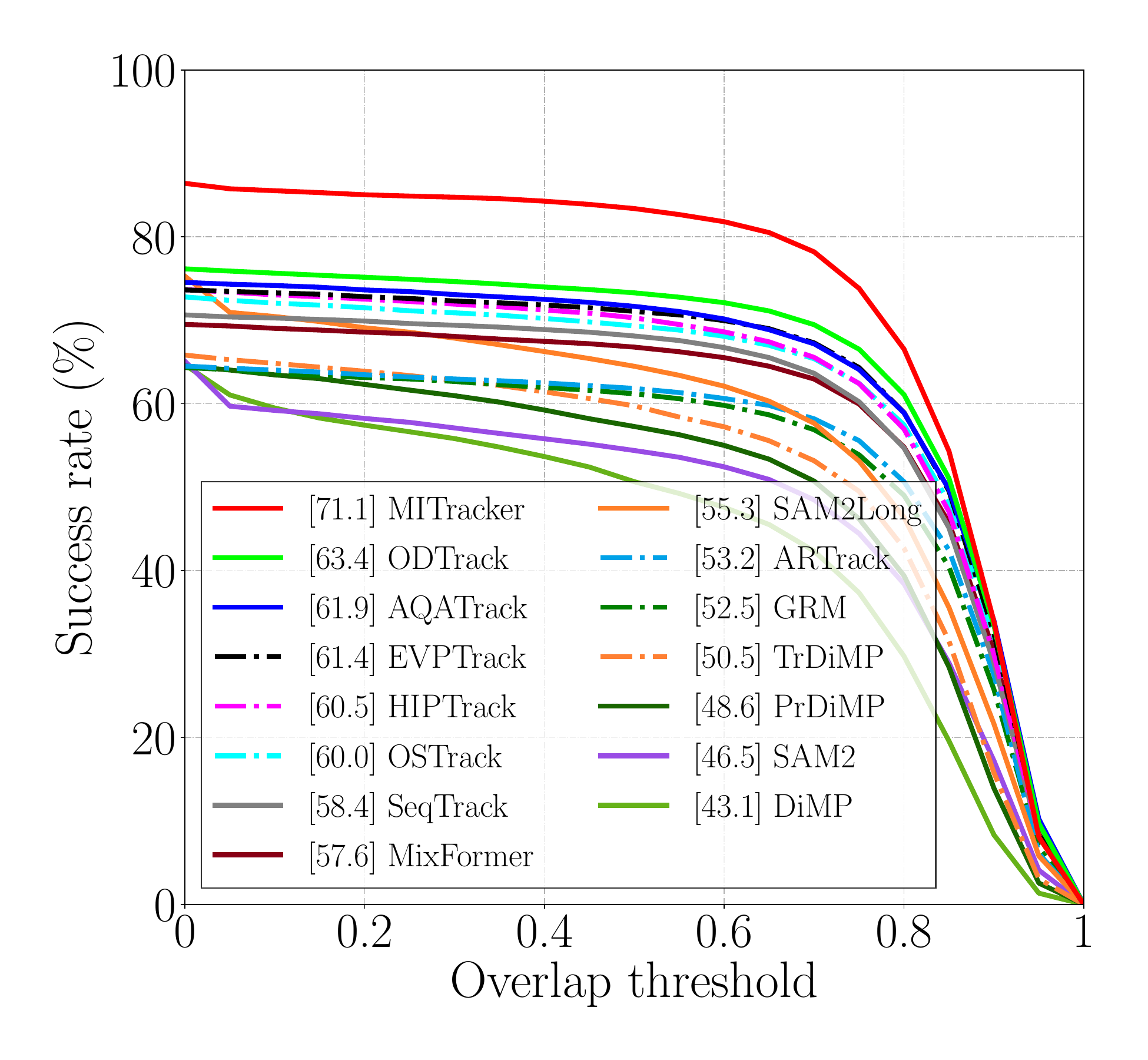}
		\caption{Success plot.}
		\label{fig: Success plot}
	\end{subfigure}
        \begin{subfigure}{0.66\columnwidth}
		\centering
		\includegraphics[width=\columnwidth]{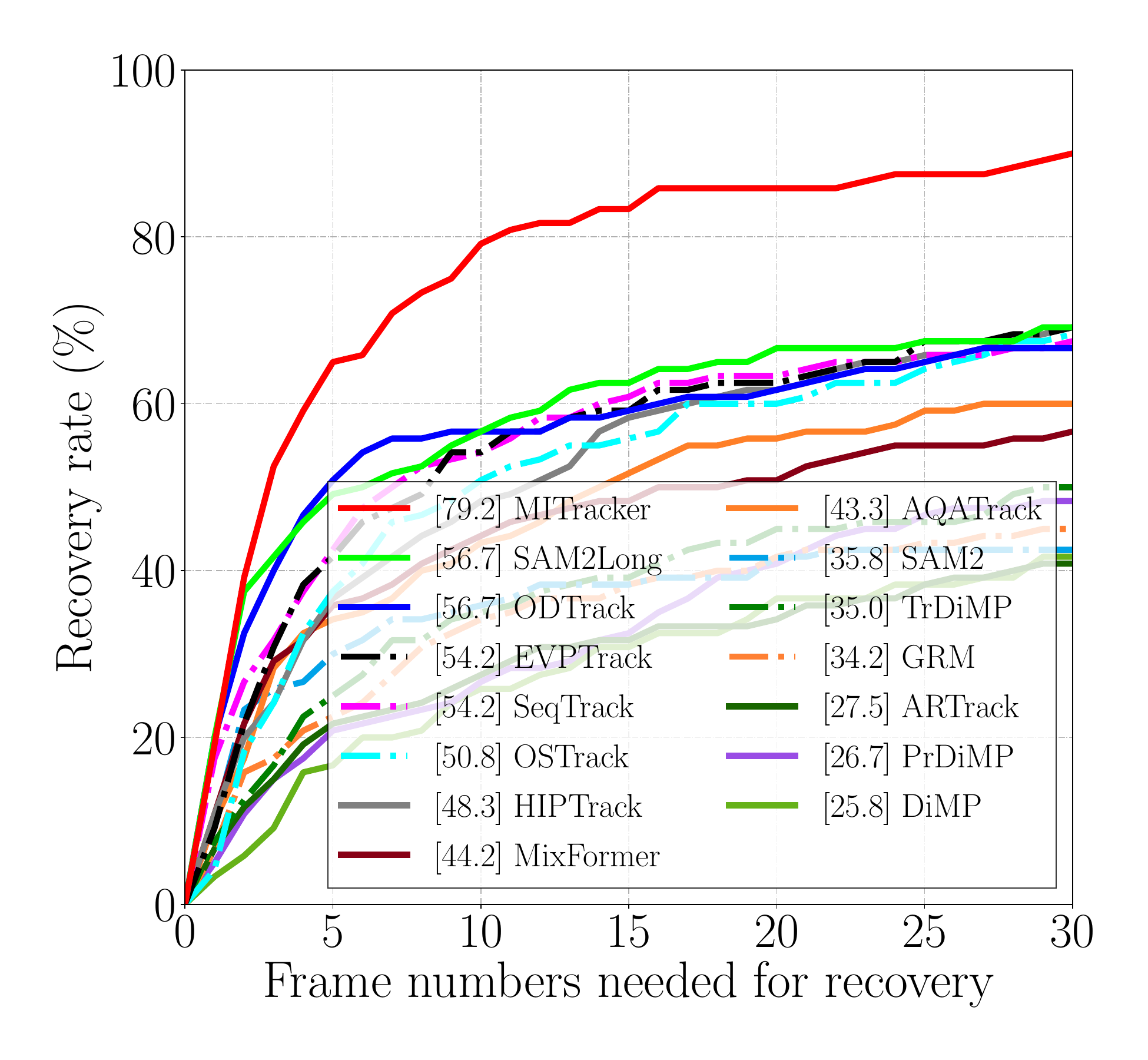}
		\caption{Recovery ability plot.}
		\label{fig: recovery_plot}
        \end{subfigure}
        \centering
        \begin{subfigure}{0.65\columnwidth}
		\centering
		\includegraphics[width=\columnwidth]{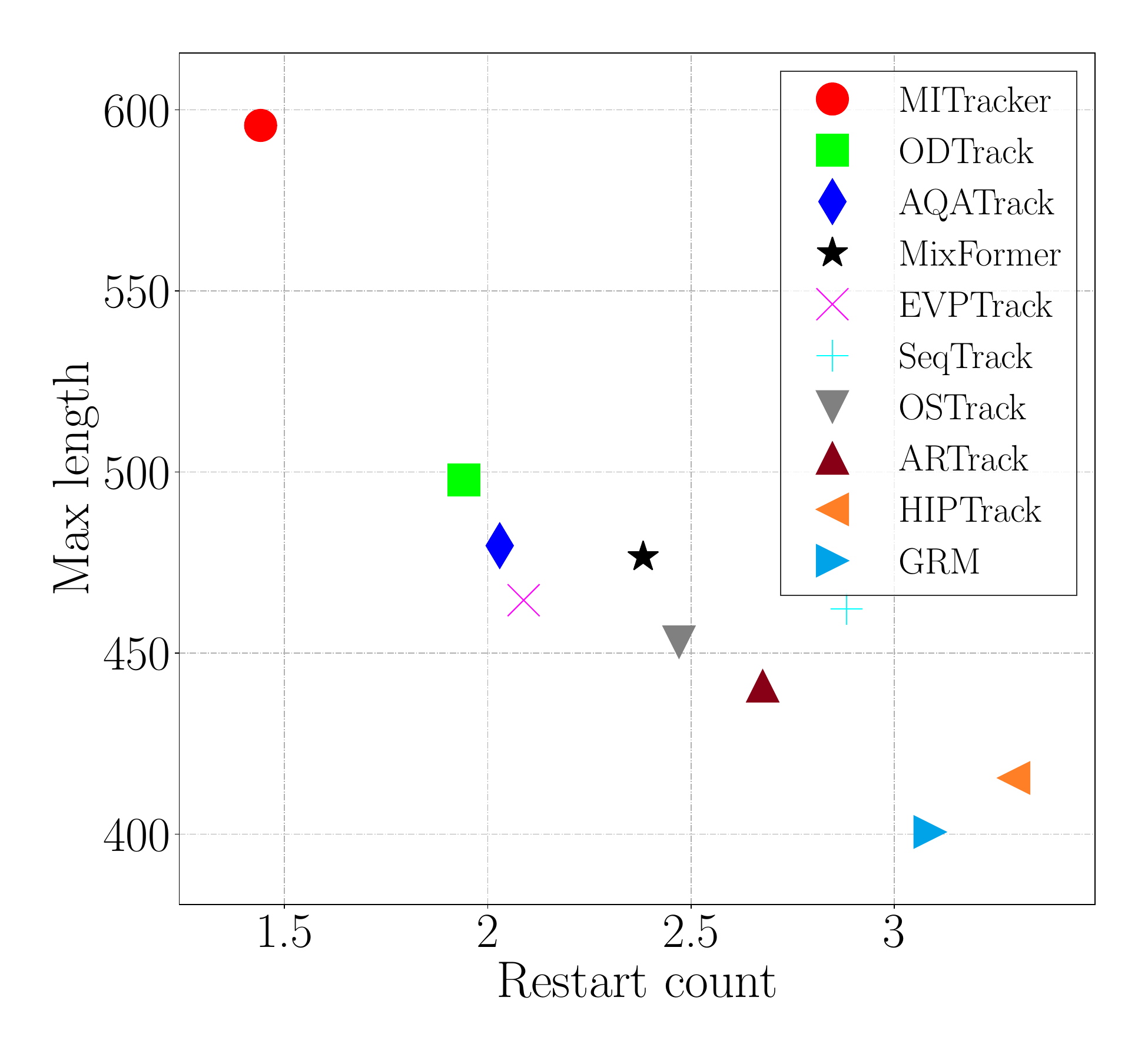}
		\caption{Robust tracking plot.}
		\label{fig: Robust tracking plot}
        \end{subfigure}
        
	\caption{General experiments on the MVTrack  dataset evaluate tracking robustness. MITracker provides multi-view results, while other methods yield single-view results. In (a), methods are ranked by AUC and noted in the legend. For (b), the numbers in the legend represent the method's recovery rate within 10 frames after the target disappears.
 }
	\label{OPE}
\end{figure*}

\textbf{Training Setup.}
We initialize our view-specific encoder with pre-trained DINOv2 \cite{oquab2023dinov2} parameters using the ViT-base model \cite{dosovitskiy2020image}. 
For the visual inputs, we set the reference frame with $182 \times 182$ pixels, and the search frame with $364 \times 364$ pixels. 
We utilize the camera parameters from the dataset for projection, with the 3D feature volume having dimensions \( X = 200 \), \( Y = 200 \), and \( Z = 3 \).

Our training process consists of two stages. In the first stage, we only train the view-specific feature extraction module.
Specifically, we train the view-specific encoder and BBox head using single-view inputs from GOT-10K and MVTrack datasets until convergence.
For each viewpoint, we include one reference frame and two random search frames from 200 frame interval in each iteration, thus promoting temporal information propagation between frames.
In the second stage, we fine-tune the view-specific encoder and train the entire framework using multi-view data from the MVTrack dataset.
For each training sample, we randomly select 2 to 4 viewpoints with one reference and two search frames in each iteration. 
All the training procedures are conducted on 2 NVIDIA A100 80GB GPUs.

For detailed implementation and training specifics, please refer to the Appendix.

\subsection{Evaluation Metrics}
We evaluate our method using three standard performance measures from the single-view tracking benchmark \cite{OTB2015, muller2018trackingnet, fan2019lasot}: Area Under Curve (AUC), Precision (P), and Normalized Precision (P$_{Norm}$):

\begin{itemize} 
    \item \textbf{AUC}: The Intersection over Union (IoU) measures the overlap between predicted and ground truth BBoxes in each frame. The AUC metric is calculated by varying the IoU threshold to evaluate the area error in the tracking region. 
    \item \textbf{P}: Precision is defined as the distance between the predicted and ground truth BBox centers. This metric is used to assess the positional error in tracking. 
    \item \textbf{P$_{Norm}$}: To mitigate biases due to variations in BBox size, we normalize the center point by the width and height of the ground truth BBox. This adjustment provides a more accurate metric. 
\end{itemize}

\subsection{Comparison with Existing Methods}
\textbf{SOTA Performance on Benchmark.}
We evaluate tracking performance with single-view visual object tracking methods, training all models (except SAM2 and SAM2Long) on the GOT10K and MVTrack datasets. The models are tested on both the MVTrack and GMTD datasets under single-view and multi-view settings.

However, single-view methods cannot handle multi-view inputs or generate multi-view predictions. To address this, we employ a post-fusion strategy to obtain multi-view results. Specifically, single-view predictions are first projected into the 3D world coordinate system. The region with maximum overlap is identified as the target position, which is then reprojected onto the 2D image plane of each viewpoint to generate the fused multi-view tracking results.

As shown in Table \ref{tab:compairsion}, MITracker achieves superior performance in both multi- and single-view tracking across different datasets. 
In multi-view scenarios with 3-4 cameras, MITracker outperforms other methods that rely on post-processing for multi-view fusion, surpassing the second-best method OSTrack by approximately 26\% in P$_{Norm}$. In single-view settings, MITracker surpasses SOTA methods on the MVTrack dataset, achieving an AUC of 68.57\%, which outperforms ODTrack by approximately 5\%.

Notably, MITracker exhibits strong generalization capabilities by achieving exceptional performance on the GMTD, despite it not being included in the training data.
This demonstrates the robustness of our multi-view approach even in single-view scenarios. We attribute these improvements to our multi-view training strategy, which enables the model to better understand spatial relationships crucial for precise tracking.
It is also noteworthy that post-processing degrades the performance of all single-view methods.
This indicates a substantial distribution gap in view-independent feature detection across models, making effective fusion through geometric projections challenging.

\textbf{Stable Continuous Tracking Capability.}
To further evaluate tracking robustness, we conducted three comparative experiments on the MVTrack dataset. Only MITracker utilizes multi-view inputs, other methods use single-view inputs and generate BBoxes independently.

First, we analyzed tracking success rates across various IoU thresholds, as shown in Figure \ref{fig: Success plot}. MITracker consistently outperforms competing methods regardless of the threshold value.

Second, we evaluated the recovery capability after the target was invisible by measuring the proportion of successful tracking resumption within given frame intervals \cite{huang2024rtracker}.
As illustrated in Figure \ref{fig: recovery_plot}, with a 10-frame interval, MITracker achieves a high success rate of 79.2\% in these recovery tests.
In comparison, SAM2Long only achieves a 56.7\% recovery rate under the same setting, highlighting our method's exceptional ability to quickly reestablish tracking after the target dissapears.

In practical applications, users can manually intervene to restart the model’s tracking by providing an accurate initial position. 
In this experimental setup, we measured the maximum continuous tracking length of video frames and the average number of restarts (triggered when target loss exceeds 10 frames, using ground truth for repositioning) \cite{zhao2024biodrone}.
As shown in Figure \ref{fig: Robust tracking plot}, MITracker achieves nearly 100 frames longer tracking duration than ODTrack while requiring fewer restart counts.

\subsection{Ablation Study}
Results in Table \ref{tab:ablation_study} demonstrate that BEV Loss, which provides implicit multi-view information feedback, significantly enhances model performance. 
This improvement is attributed to its ability to augment spatial awareness during single-view feature extraction. 
The Spatial attention, which utilizes fused information to adjust outputs from single-view perspectives, also contributes to notable performance improvements in the model.

\input{tables/experiments/ablation}

\begin{figure}[t]
\centering
\includegraphics[width=1.0\columnwidth]{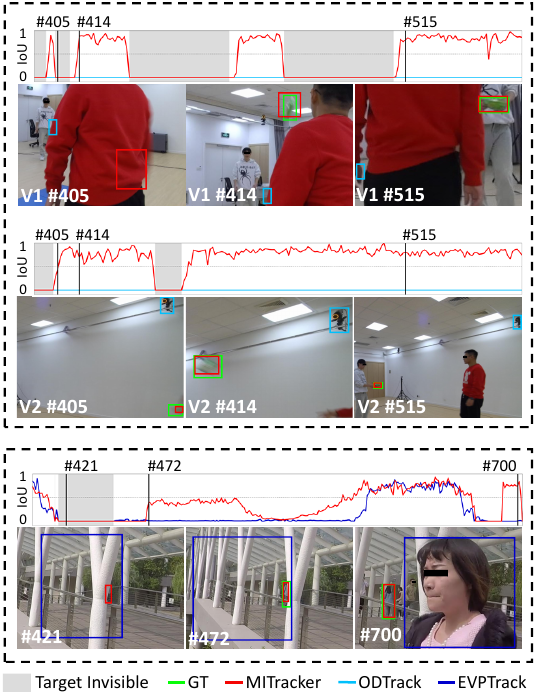} 
\caption{Qualitative comparison results. Comparison of our tracker with two SOTA methods on MVTrack dataset (top) and GMTD (bottom). Each frame is cropped for better visualization. IoU curves of each method’s prediction and ground truth are shown above, where IoU reflects tracking quality. MITracker demonstrates superior re-tracking performance upon target reappearance.
}
\label{visulazation}
\end{figure}
\subsection{Visualization Comparison}

Our qualitative evaluation focuses on the influence of occlusion and fast motion. In the upper part of Figure \ref{visulazation}, we select two viewpoints from the MVTrack dataset and evaluate them on MITracker and ODTrack, which has the second-best performance on this dataset. 
The \colorbox{mygray}{gray areas} in the graph represent periods when the object is out of view or fully occluded by other objects. 
We can easily observe that MITracker is able to re-track the object shortly after it reappears, whereas ODTrack tends to continue in a lost state. 
For instances \#405 and \#515 in V2, even when the object reappears in the frame, ODTrack still mistakenly locks onto the wrong object. 
The bottom of Figure \ref{visulazation} presents tests conducted on the GMTD, where we also selected the second-best method, EVPTrack, for comparison with MITracker. 
When a pedestrian reappears after being obscured by a pillar, EVPTrack mistakenly locks onto the wrong target, whereas MITracker is able to maintain stable and continuous tracking.

\begin{figure}[t]
\centering
\includegraphics[width=1.0\columnwidth]{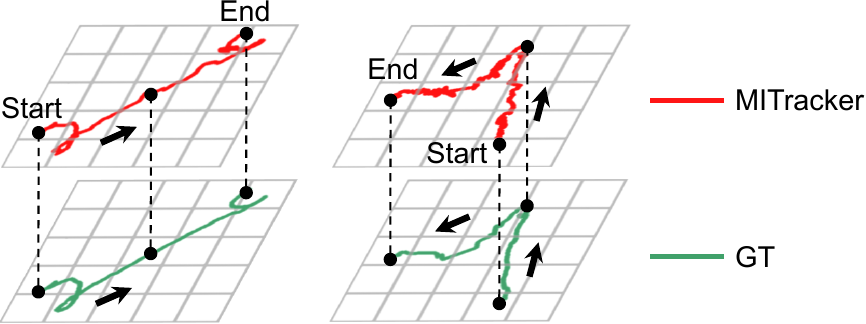} 
\caption{Visualization of BEV trajectories on the MVTrack dataset. Left: scene captured by four cameras. Right: scene captured by three cameras.}
\label{fig: trajectory}
\end{figure}

We also visualize the predicted BEV trajectories from MITracker in Figure \ref{fig: trajectory}. 
Referencing the ground-truth trajectories, MITracker effectively integrates multi-view features and provides accurate 3D spatial information.

\section{Discussion}
In previous sections, we provide a detailed introduction to the MVTrack dataset and demonstrate the outstanding performance of MITracker. However, there are some areas that could be improved in future work.

\textbf{Limitations.} Although MVTrack dataset includes a diverse set of scenes, it currently consists of indoor environments only, potentially limiting the generalization of methods trained on it to outdoor settings. Additionally, MITracker relies on camera calibration for multi-view fusion, which may restrict its applicability in scenarios where calibration is challenging or infeasible.

\textbf{Future work.} We plan to extend MVTrack dataset by including outdoor scenes and a wider range of tracking objects to enable the development of more generalizable multi-view tracking algorithms. Furthermore, we aim to enhance MITracker by reducing its dependency on precise camera calibration, making it more adaptable to scenarios where accurate calibration is challenging.

%% file: tables/experiments/compairson.tex
\begin{table*}[ht]
\centering
% \resizebox{\linewidth}{!}{
\begin{tabular}{l ccc  ccc ccc}
\toprule
Dataset & \multicolumn{6}{c}{MVTrack} & \multicolumn{3}{c}{GMTD}\\ \hline
\multirow{2}{*}{Method} & \multicolumn{3}{c}{Single-View} & \multicolumn{3}{c}{Multi-View} & \multicolumn{3}{c}{Single-View}\\
\cmidrule[0.5pt](lr){2-4}
\cmidrule[0.5pt](lr){5-7}
\cmidrule[0.5pt](lr){8-10}
 & AUC(\%) & P$_{Norm}$(\%)  & P(\%) & AUC(\%) & P$_{Norm}$(\%)  & P(\%) & AUC(\%) & P$_{Norm}$(\%)  & P(\%) \\ 
\hline
DiMP \cite{bhat2019learning}& 43.14 & 59.52 & 53.13 & 35.77& 49.04& 51.65& 52.71& 68.24&66.04\\% 1325
PrDiMP \cite{danelljan2020probabilistic}& 48.61 & 66.09 & 58.93 & 38.49& 54.68& 57.95& 57.76& 76.21&70.49\\% 661
TrDiMP \cite{wang2021transformer}& 50.54 & 67.67 & 60.44 & 39.71& 55.31& 58.52& 59.51& 78.94&73.48\\% 698
MixFormer \cite{cui2022mixformer}& 57.59& 75.44& 67.72& 43.29 & 58.07 & 62.70 & 62.03& 82.60&78.48\\% 487
OSTrack \cite{ye2022joint}& 60.04& 77.72& 70.06& \underline{49.10} & \underline{65.19} & 67.34 & 58.44& 77.37&73.23\\% 378
GRM \cite{gao2023generalized} & 52.53 & 69.91 & 62.31& 41.47 & 57.33 & 58.76 & 55.67&  74.02&70.27\\% 83
SeqTrack \cite{chen2023seqtrack}& 58.37& 76.63 & 69.03& 43.88 & 59.11 & 63.60 & 62.97& 83.20&79.32\\% 138
ARTrack \cite{wei2023autoregressive} & 53.23 &70.25 &62.49& 42.52 & 58.00 & 60.50 & 59.56& 78.12&74.23\\% 99
HIPTrack \cite{cai2024hiptrack}& 60.45 & 78.92& 70.53& 48.43 & 63.69& 66.26& 62.20& 80.62&76.94\\% 7
EVPTrack \cite{shi2024evptrack}& 61.37 & 79.76 & 71.97& 46.36 & 61.84 & 67.20 & \underline{63.89}& \underline{83.76}&\underline{79.93}\\% 6
AQATrack \cite{xie2024autoregressive} & 61.93 & 80.00 & 72.69& 45.24 & 59.76 & 65.33 & 63.57 & 83.04 & 79.44\\%7
ODTrack \cite{zheng2024odtrack}& \underline{63.36}& \underline{82.25}& \underline{74.46}& 48.05 & 63.55 & \underline{67.70} & 61.43& 82.37&78.35\\% 13
% DiffusionTrack \cite{luo2024diffusiontrack} & & & \\%11
SAM2$^{\ast}$ \cite{ravi2024sam2} & 46.49 & 63.12 & 56.82 & 39.08& 53.49& 57.34& 59.88& 74.66&73.25\\ 
SAM2Long$^{\ast}$ \cite{ding2024sam2long} & 55.30 & 72.84 & 67.40 & 45.40& 59.12& 65.30& 62.80&78.60&77.40\\ 
\hline
MITracker & \textbf{68.57} &  \textbf{88.77}  & \textbf{80.93}  & \textbf{71.13}& \textbf{91.87}& \textbf{83.95}& \textbf{65.96}& \textbf{87.05} & \textbf{82.07}\\
\bottomrule
\end{tabular}
% }
% \caption{Comparison with SOTA methods on the MVTrack and GMTD datasets. Method marked with ${^\ast}$ indicate the use of pre-trained weights without fine-tuning on any dataset. The best results are \textbf{bolded}, while the second-best results are \underline{underlined}.}
\caption{Comparison with SOTA methods on the MVTrack and GMTD datasets. MITracker is for multi-view tracking, while others are single-view methods. Methods with ${^\ast}$ use pre-trained weights without fine-tuning. Best results are \textbf{bolded}, second-best are \underline{underlined}.}
\label{tab:compairsion}
\end{table*}

%% file: tables/experiments/ablation.tex
\begin{table}[ht]
\centering
\resizebox{\linewidth}{!}{
\begin{tabular}{Xccccc}
\toprule
BEV Loss & Spatial Attention & AUC(\%) & P$_{Norm}$(\%) & P(\%) \\ \midrule
&   & 63.99 & 82.82& 75.00 \\
% &   & 68.23& 87.16 & 79.22 \\
\cmark &   &69.64&89.85& 82.01\\
\cmark & \cmark &71.13& 91.87& 83.95\\ \bottomrule
\end{tabular}
}
\caption{Ablation study for multi-view evaluation on MVTrack dataset.}
\label{tab:ablation_study}
\end{table}

%% file: sections/X_suppl.tex
\clearpage
\setcounter{page}{1}
\maketitlesupplementary

% In the supplementary material, we provide additional information about the MVTrack dataset in Section \ref{sec: dataset details}. Further implementation details and experimental results of MITracker are available in Section \ref{sec: experiment details}.
Section \ref{sec: dataset details} provides additional information on the MVTrack dataset, while Section \ref{sec: experiment details} includes further implementation details and experimental results of MITracker.

\section{Dataset Details}
\label{sec: dataset details}
\textbf{Data Annotation.}
In the BEV annotations, the MVTrack dataset covers an $8m \times 8m$ area. Ground truth labels are projected onto a $400 \times 400$ grid, where each cell is $2 cm \times 2 cm$ in size.
 
% 目标属性
\textbf{Attributes Definition.} MVTrack dataset contains nine attributes to assess tracking robustness, as shown in Table \ref{tab: attributes description}. We provide frame-level binary labels for five attributes: Background Clutter (BC), Motion Blur (MB), Partial Occlusion (POC), Full Occlusion (FOC), and Out of View (OV). These are manually annotated for each frame. Deformation (DEF) is labeled according to whether the tracked target deforms. Low Resolution (LR), Aspect Ratio Change (ARC), and Scale Variation (SV) are automatically computed from changes in the BBox size.

\input{tables/dataset/attributes_table}

\begin{figure}[h]
    \centering
    \begin{subfigure}[b]{0.62\columnwidth}
        \centering
        \includegraphics[width=\columnwidth]{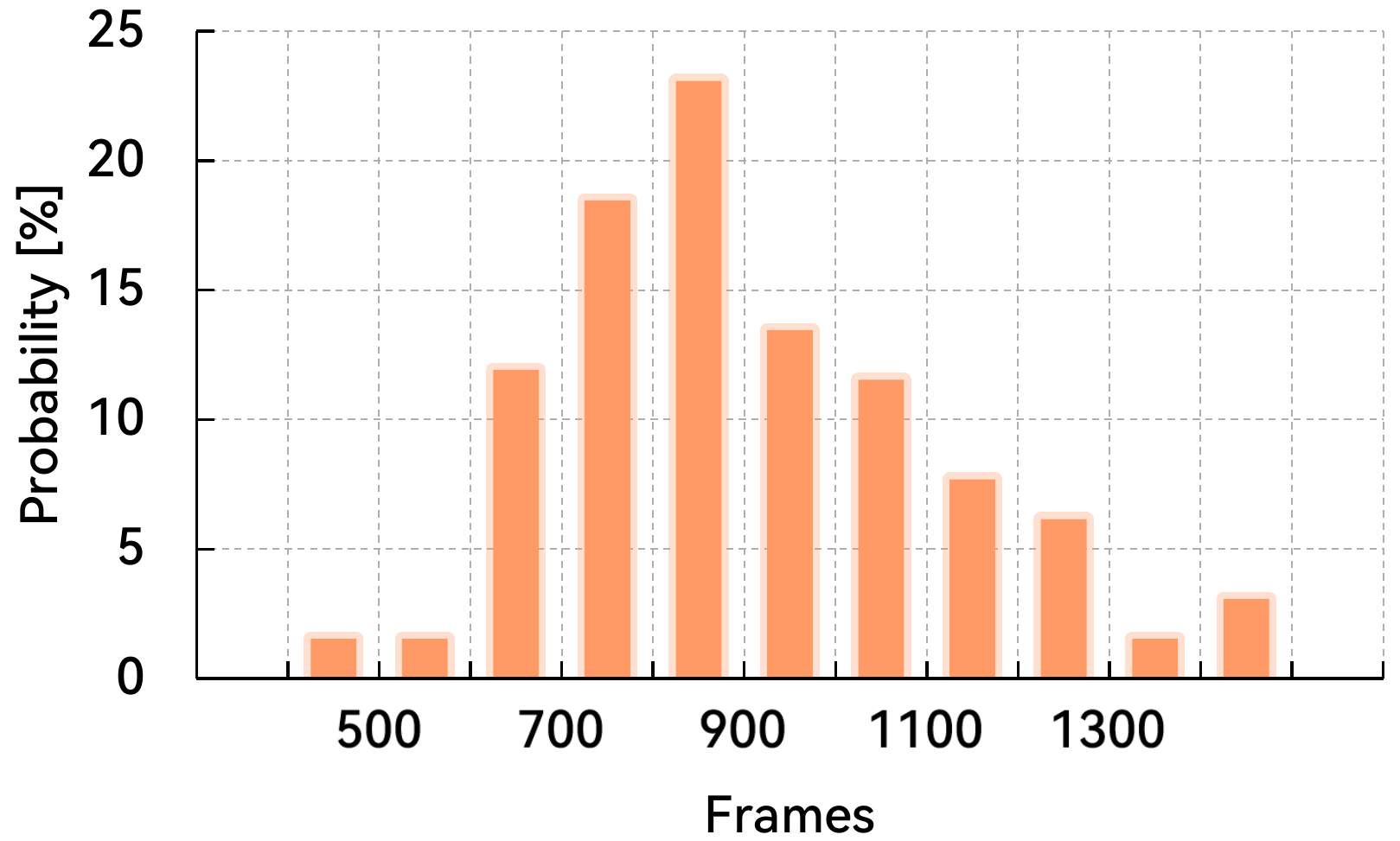}
        \caption{Frames distribution.}
        \label{fig: frames distribution}
    \end{subfigure}
    
    \begin{subfigure}[b]{0.62\columnwidth}
        \centering
        \includegraphics[width=\columnwidth]{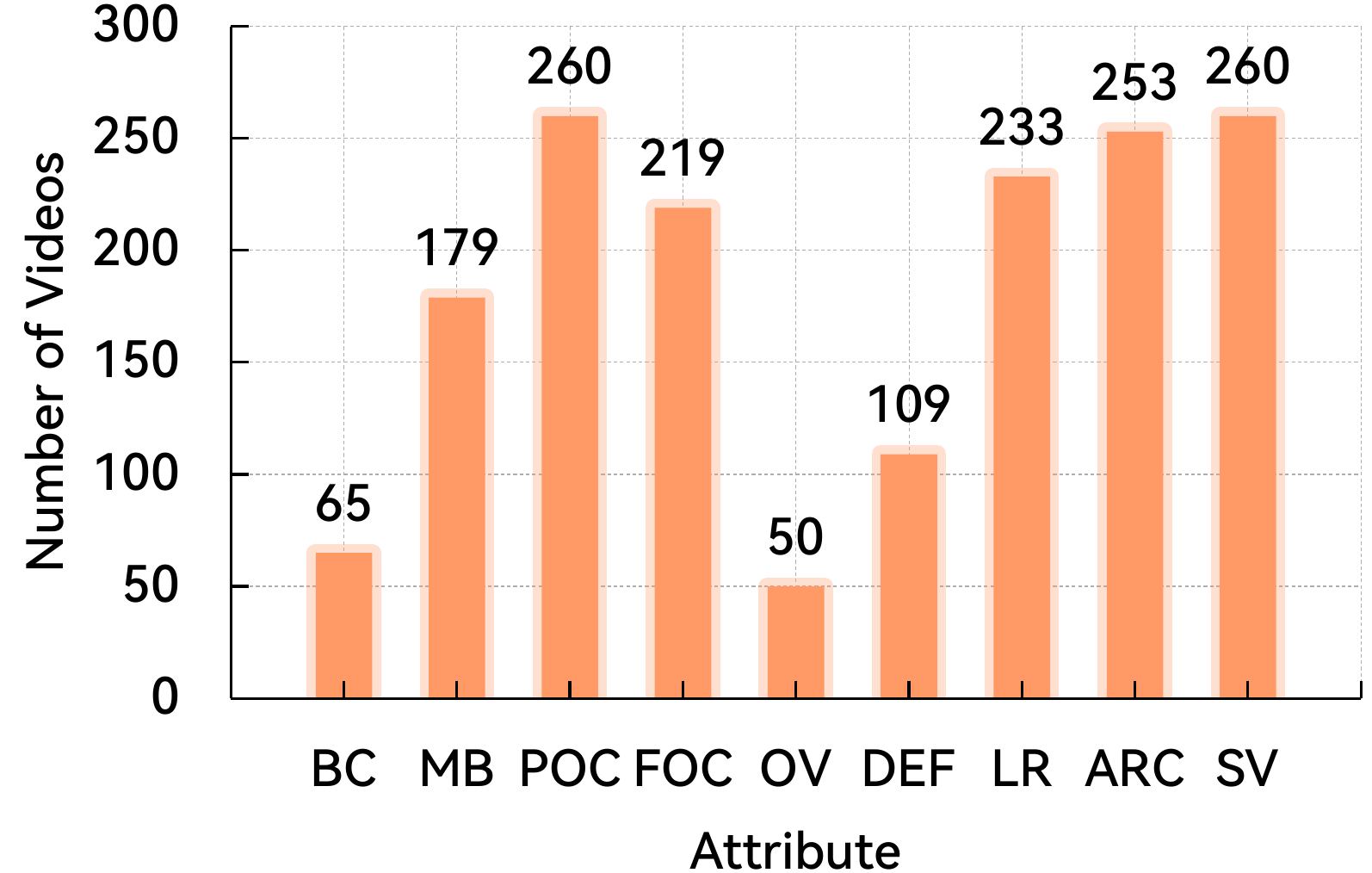}
        \caption{Attributes distribution.}
        \label{fig: attributes distribution}
    \end{subfigure}
    \caption{Distribution of sequences in each attribute and length in our MVTrack dataset.}
    \label{fig:distribution}
\end{figure}

\textbf{Statistical Details.}
The MVTrack dataset contains 260 videos averaging around 900 frames each, as shown in Figure \ref{fig: frames distribution}. As illustrated in Figure \ref{fig: attributes distribution}, a key challenge is occlusion, which often results from subject-object interactions that cause partial or complete occlusion. Consequently, tracking models need to manage occlusion to perform robustly and adeptly on this dataset.

\section{Experiment Details}
\label{sec: experiment details}

\subsection{Training and Resource Analysis}
\textbf{Training Details.} 
We process the visual inputs by cropping the reference frame to 2 times the target's BBox size and resizing it to $182 \times 182$ pixels. The search frame is cropped at 4.5 times the target box area and resized to $364 \times 364$ pixels to expand the search region. During projection, we transform the camera intrinsic matrix $C_K$ accordingly and add noise to the translation vector $C_t$ to prevent overfitting in multi-view fusion.

Training consists of two stages. In the first stage, we optimize the view-specific encoder using AdamW with a learning rate of $1 \times 10^{-5}$ and the rest of the model at $1 \times 10^{-4}$. We train for 50 epochs, sampling 10,000 image pairs per epoch with a batch size of 32. In the second stage, we fine-tune the encoder at $1 \times 10^{-6}$ while keeping other components at $1 \times 10^{-4}$. We use the MVTrack dataset, sampling 2,500 multi-view image pairs per epoch for 40 epochs with a batch size of 4. AdamW is used throughout.

\textbf{Computational Resource.}
We evaluate MITracker and the single-view model ODTrack under the same input (4 views) on an NVIDIA A100, as summarized in Table \ref{tab:time}. Although multi-view fusion introduces additional computational overhead, it remains within an acceptable range.
\begin{table}[h]
\centering
\begin{tabular}{lccc}
\toprule
Method & Parameters~(M) & GRAM~(MB) & FPS \\ \midrule
ODTrack & 92.12 & 365.82& 18.78 \\
MITracker &101.65& 407.78& 14.08\\ \bottomrule
\end{tabular}
\caption{Comparison of computational complexity and resource.}
\label{tab:time}
\end{table}

\begin{figure*}[th]
	\centering
	\begin{subfigure}{0.8\columnwidth}
		\centering
		\includegraphics[width=\columnwidth]{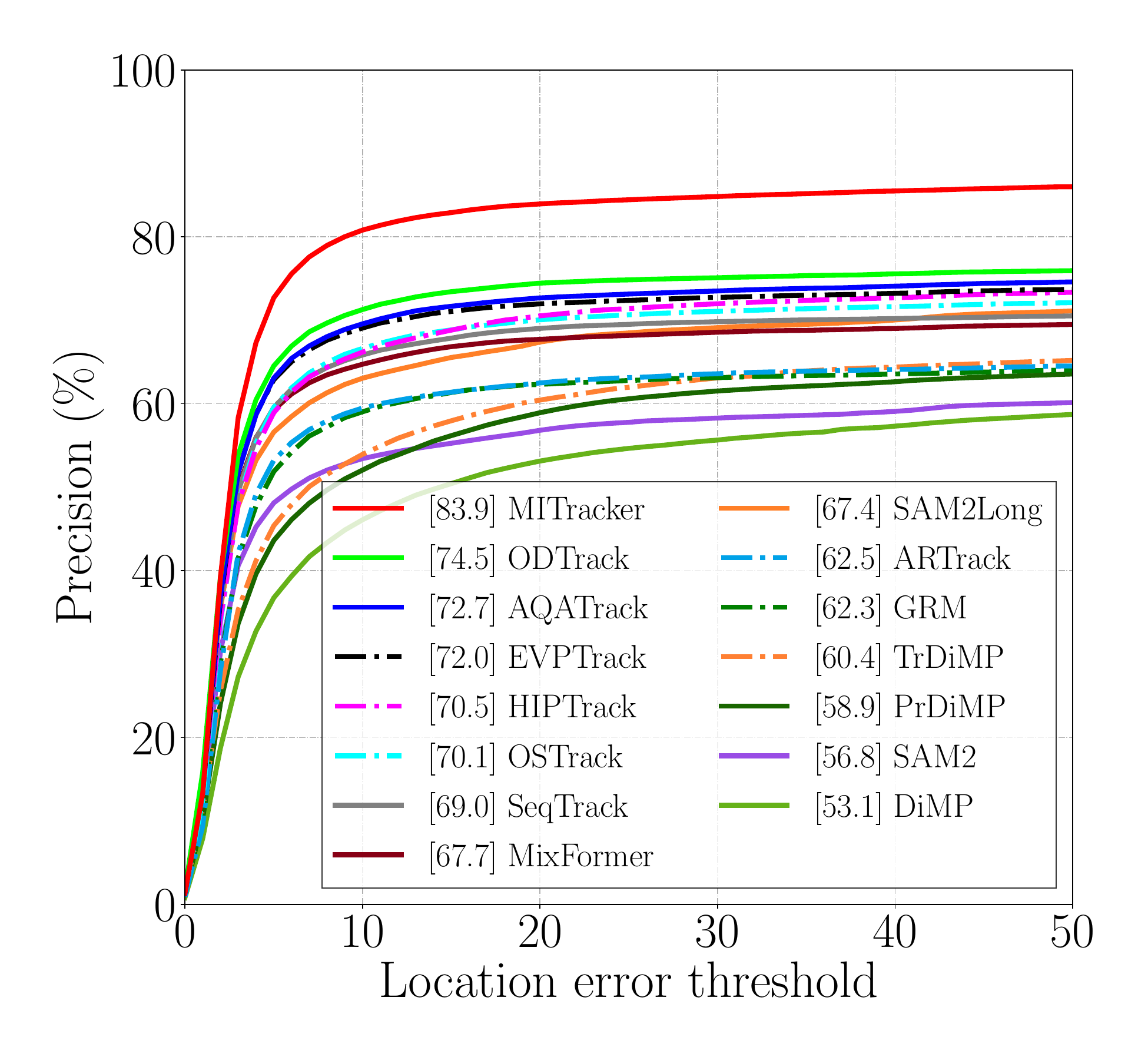}
		\caption{Precision plot on MVTrack dataset.}
		\label{fig: Precision plot mvtrack}
	\end{subfigure}
        \hspace{5mm}
        \begin{subfigure}{0.8\columnwidth}
		\centering
		\includegraphics[width=\columnwidth]{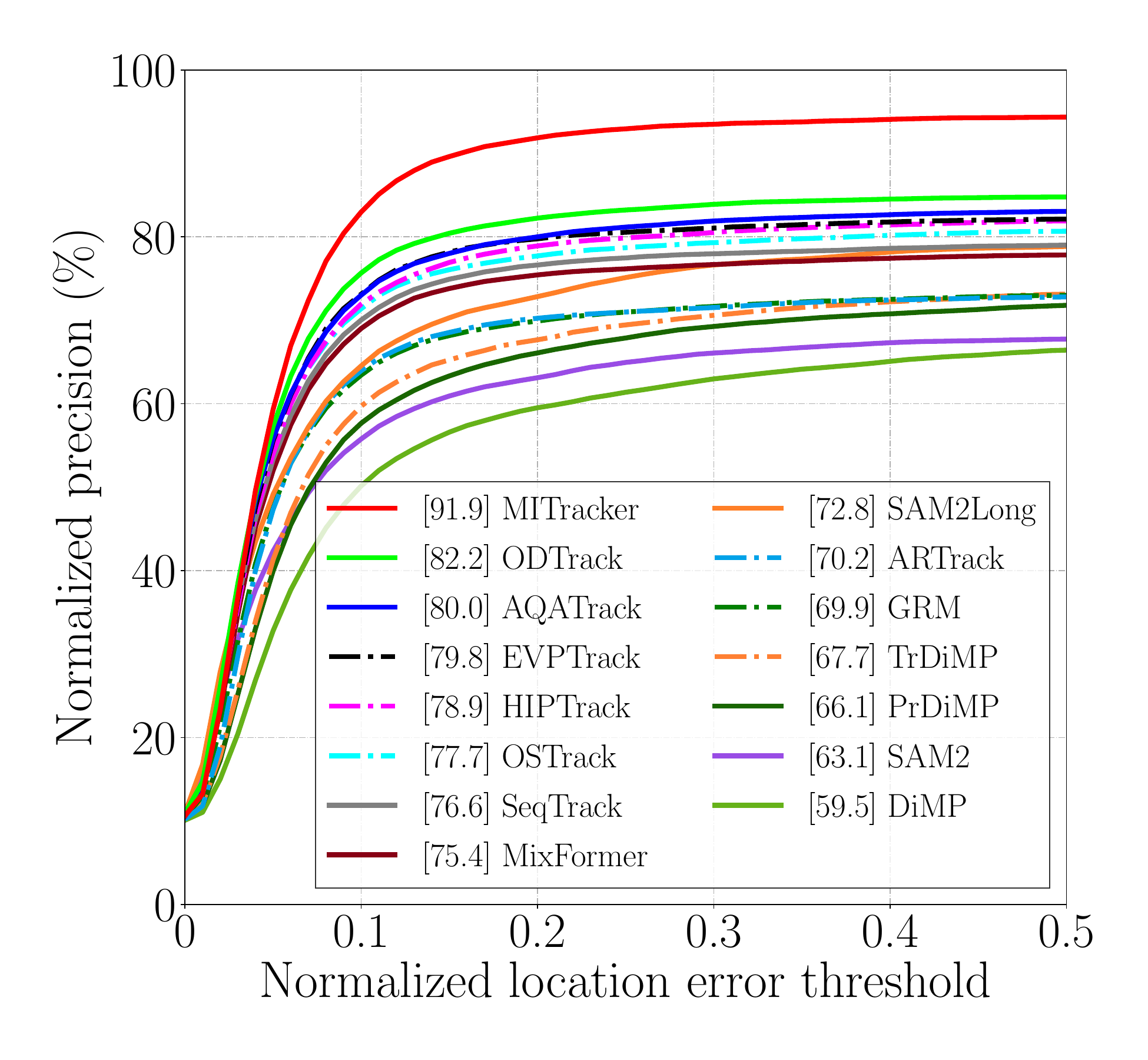}
		\caption{Normalized precision plot on MVTrack dataset.}
		\label{fig: normalized Precision plot mvtrack}
        \end{subfigure}

        \vspace{5mm}
        \centering
        \begin{subfigure}{0.8\columnwidth}
		\centering
		\includegraphics[width=\columnwidth]{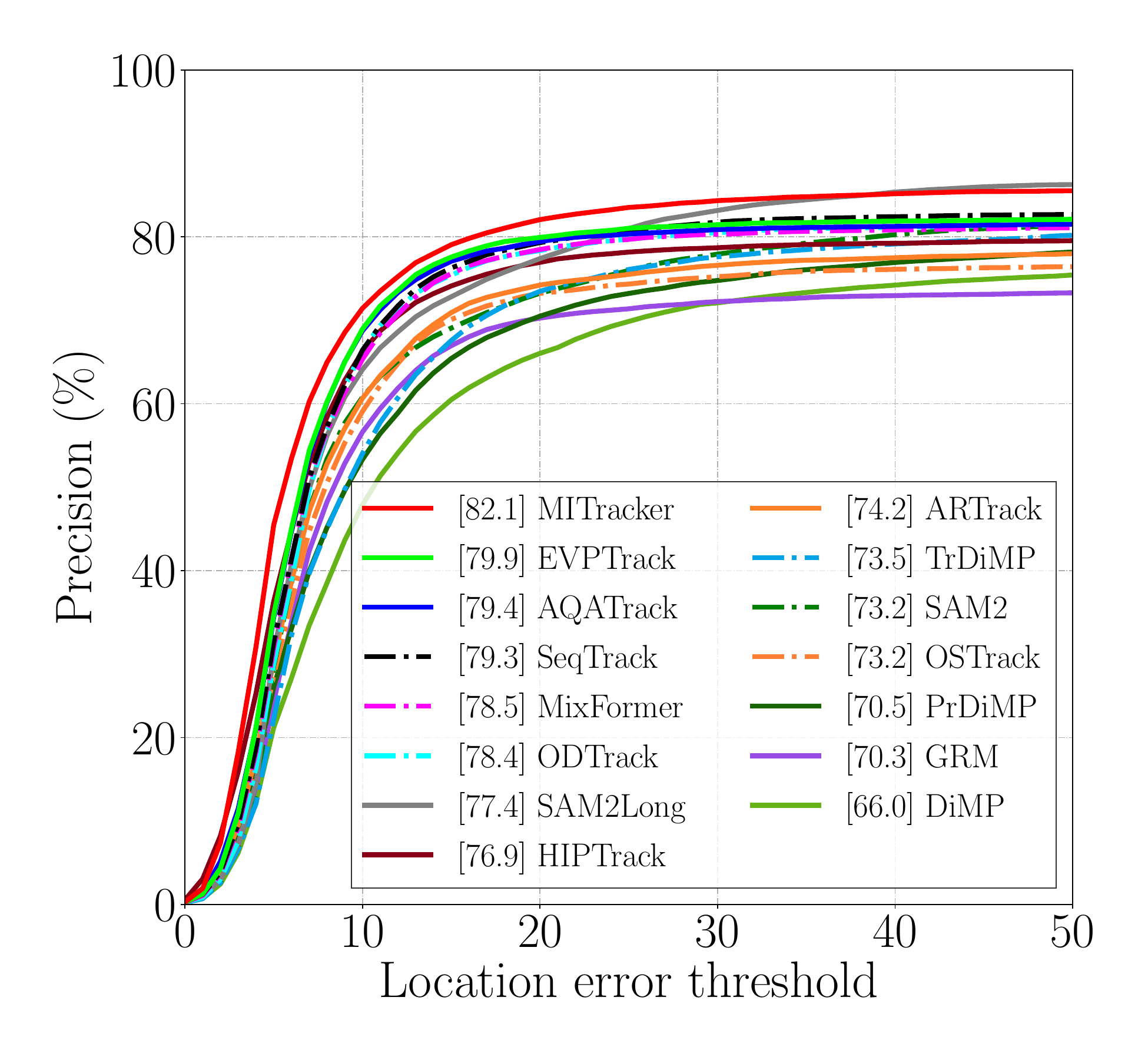}
		\caption{Precision plot on GMTD.}
		\label{fig: precision plot gmtd}
        \end{subfigure}
        \hspace{5mm}
        \centering
        \begin{subfigure}{0.8\columnwidth}
		\centering
		\includegraphics[width=\columnwidth]{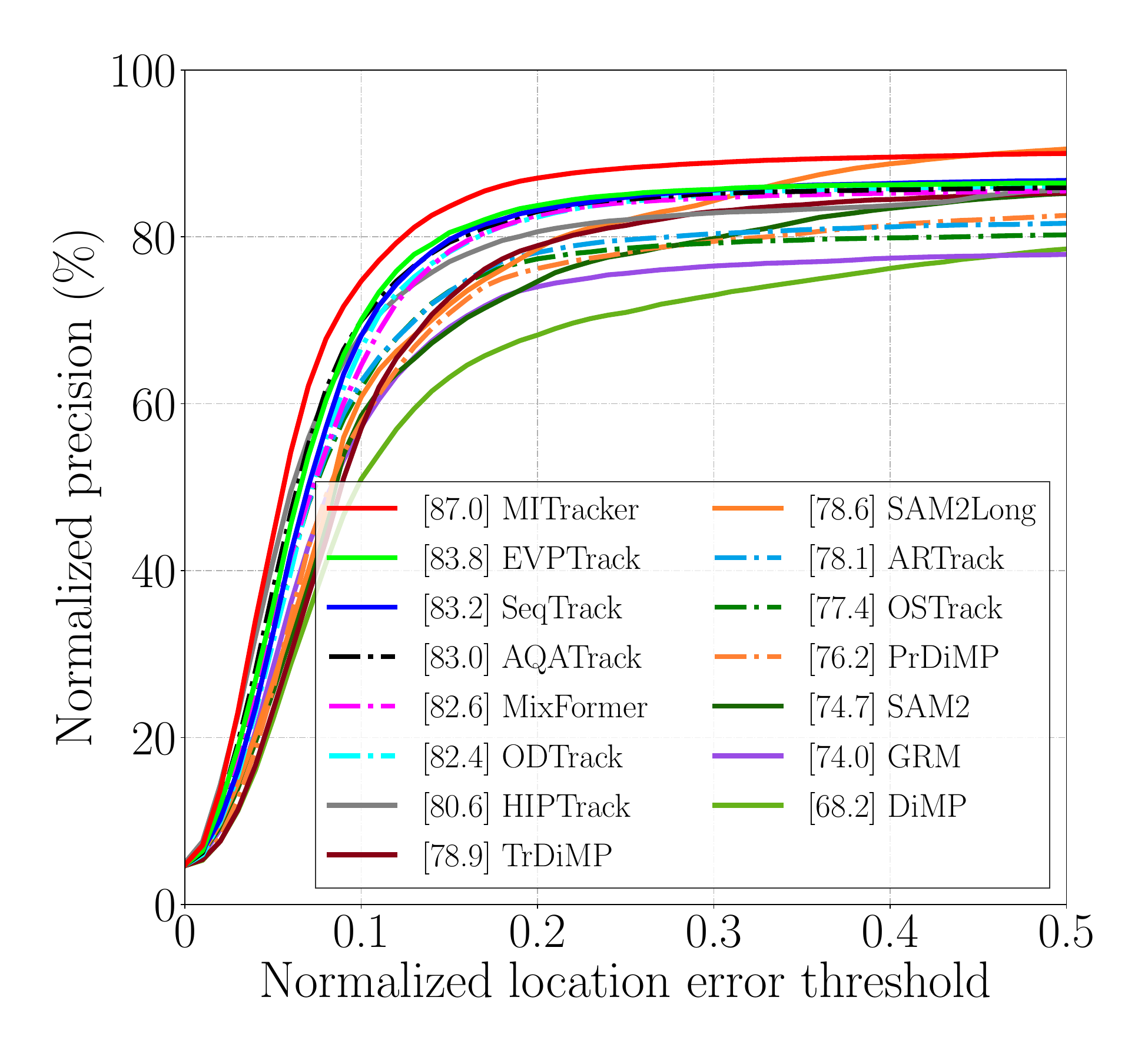}
		\caption{Normalized precision plot on GMTD.}
		\label{fig: normalized precision plot gmtd}
        \end{subfigure}

        \vspace{5mm}
        \centering
        \begin{subfigure}{0.8\columnwidth}
		\centering
		\includegraphics[width=\columnwidth]{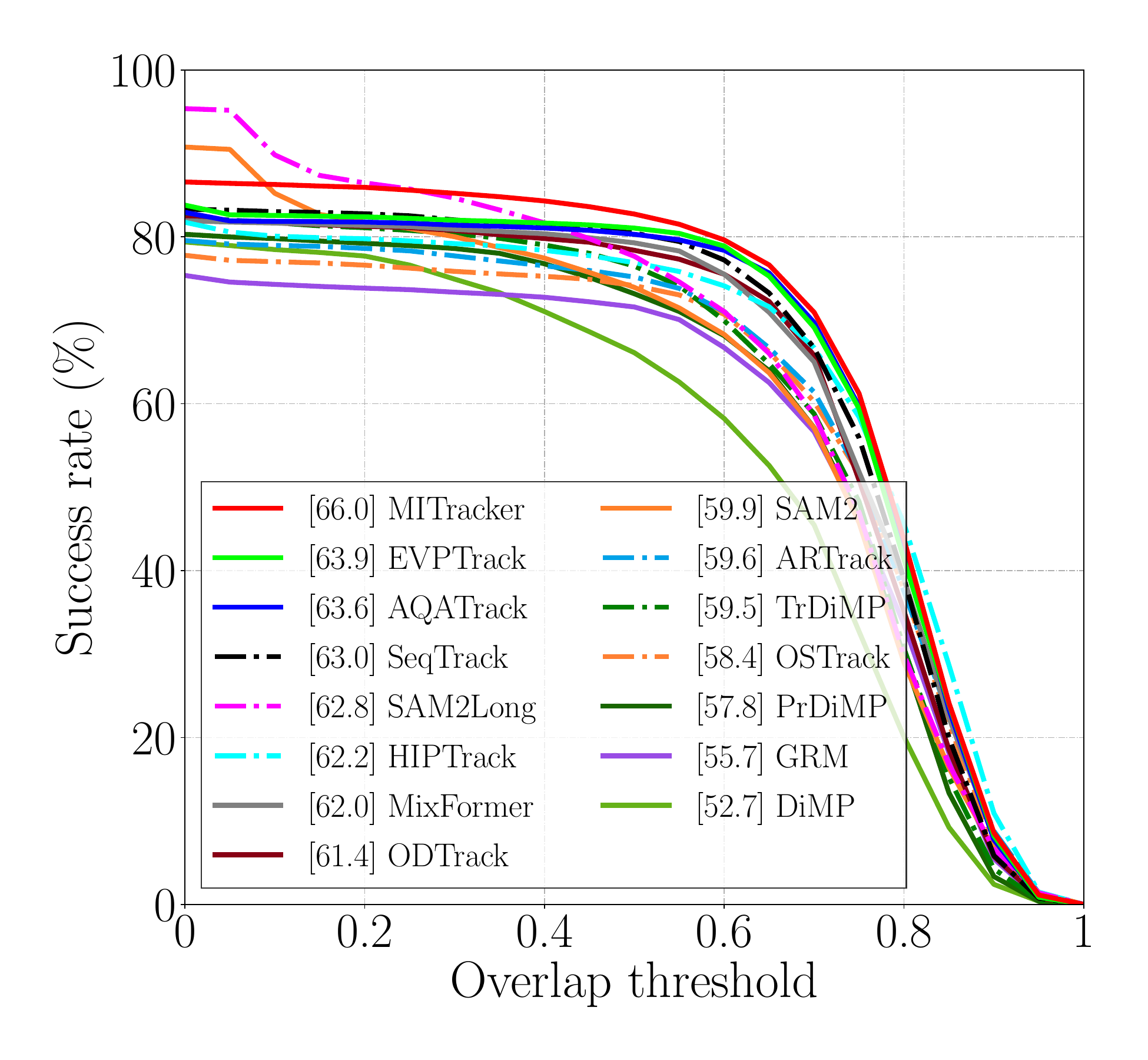}
		\caption{Success plot on GMTD.}
		\label{fig: Success plot gmtd}
        \end{subfigure}
        \hspace{5mm}
        \centering
        \begin{subfigure}{0.785\columnwidth}
		\centering
		\includegraphics[width=\columnwidth]{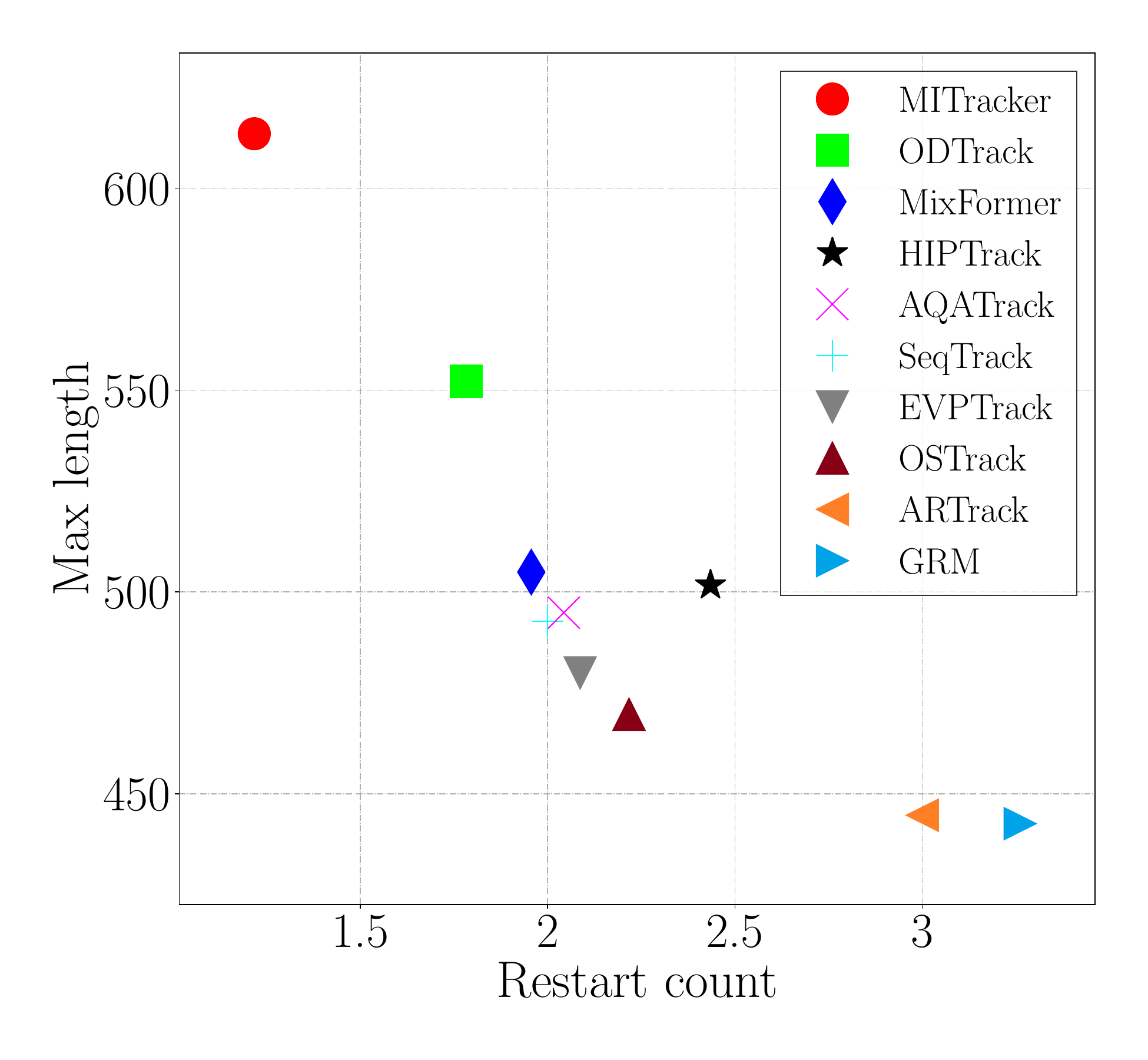}
		\caption{Robust tracking plot on GMTD.}
		\label{fig:  Robust tracking plot gmtd}
        \end{subfigure}
        
	\caption{Comparative results across MVTrack and GMTD datasets, with rankings noted in the legends. Parts (a) and (c) sort methods by P with a 20-pixel threshold, parts (b) and (d) by P$_{norm}$ with a 0.2 threshold, and part (e) by AUC.}
	\label{OPEmore}
\end{figure*}

\begin{figure*}[t]
\centering
\includegraphics[width=0.9\textwidth]{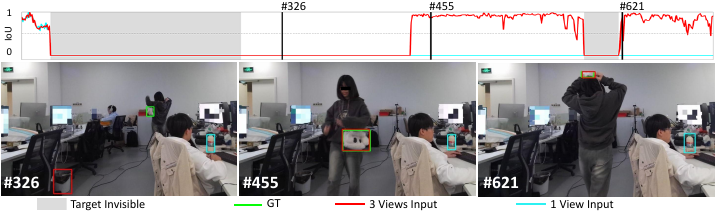} 
\caption{Qualitative comparison results on the impact of different numbers of input views. For a specific view, we compare the effects of using only that view versus including two additional overlapping views.}
\label{fig: input view}
\end{figure*}

\subsection{Comparison on Benchmark Details}
In Figure \ref{OPEmore}, we provide further quantitative evaluations of the AUC, P, and P$_{norm}$ across various threshold settings for both the MVTrack and GMTD datasets. In most settings, MITracker consistently outperforms other methods. 

During zero-shot testing on the GMTD, SAM2 and SAM2Long perform better under lenient threshold conditions but lacks the ability to localize objects precisely.
Furthermore, as shown in Figure \ref{fig: Robust tracking plot gmtd}, MITracker sustains longer tracking durations with fewer reinitializations on this unseen dataset.

\subsection{More Ablation Study}
\textbf{Impact of Input Views.} To assess the importance of the number of views for tracking, we select a fixed camera from each scenario in the testing set. We then examine how model performance changes as we increase the number of additional cameras. The results in Table \ref{tab: input view} highlight the benefits of adding more cameras.

Figure \ref{fig: input view} illustrates the challenges faced by the single-view model: after a prolonged target disappearance, it mistracks a white bottle. In contrast, the multi-view model initially mistakes a white trash can for the target but quickly recovers and maintains stable tracking with the aid of additional views.

\begin{table}[h]
\centering
\resizebox{0.8\linewidth}{!}{
\begin{tabular}{Xcccc}
\toprule
Input views & AUC(\%) & P$_{Norm}$(\%) & P(\%) \\  \hline
1 & 62.27 & 84.71& 73.92 \\
2 & 63.97& 87.07 & 76.30 \\
3 & 67.97& 91.50 & 80.73\\
3/4 &68.65& 92.37&81.55\\ \bottomrule
\end{tabular}
}
\caption{Ablation study for the impact of different numbers of input views on MVTrack dataset.}
\label{tab: input view}
\end{table}

\textbf{Impact of Multi-View Training.} 
Our experiment shows that multi-view training improves single-view performance by exposing the model to richer spatial information, which enhances its ability to handle occlusion and reappearance. Table \ref{tab:single-view} compares results with MITracker SV trained under single-view settings, highlighting the advantages of multi-view training even for single-view scenarios.

\begin{table}[h]
\centering
\begin{tabular}{l ccc}
\toprule
Method &  AUC (\%) & P$_{Norm}$ (\%)  & P (\%) \\ \hline
MITracker SV& 63.42& 82.97 & 79.67\\  % 
MITracker & 65.96& 87.05 & 82.07\\
\bottomrule
\end{tabular}
\caption{Zero-shot performance of single-view results on GMTD.}
\label{tab:single-view}
\end{table}

\textbf{Impact of Temporal Token.}
The temporal token incorporates tracking information from previous frames, Table \ref{tab:temporaltoken} highlights the improvements achieved through the temporal token.

\begin{table}[h]
\centering
\begin{tabular}{cccc}
\toprule
Temporal Token & AUC (\%) & P$_{Norm}$ (\%) & P (\%) \\ \midrule
 & 69.30 & 89.62 & 81.60 \\
\cmark &71.13& 91.87& 83.95\\ \bottomrule
\end{tabular}
\caption{Ablation study for temporal token.}
\label{tab:temporaltoken}
\end{table}

\subsection{More Visualization Results}
\begin{figure*}[th]
	\centering
	\begin{subfigure}{0.9\textwidth}
		\centering
		\includegraphics[width=\textwidth]{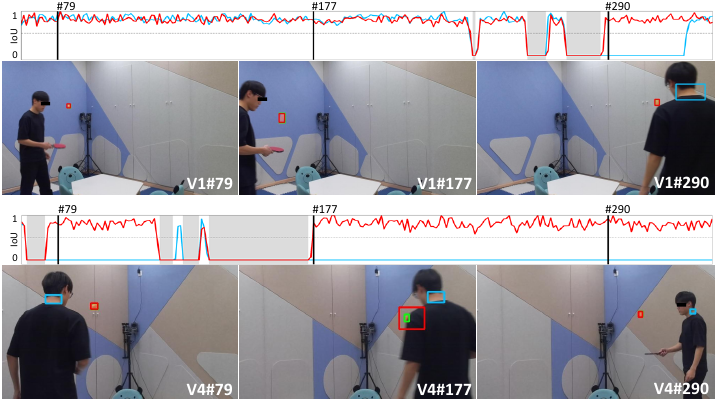}
		\caption{Two views: \textit{pingpong5-1} and \textit{pingpong5-4}. ODTrack tends to lose track after extended periods of target disappearance, whereas MITracker demonstrates robust recovery capabilities.}
		\label{fig: pingpong5}
	\end{subfigure}

        \vspace{3mm}
        \begin{subfigure}{0.9\textwidth}
		\centering
		\includegraphics[width=\textwidth]{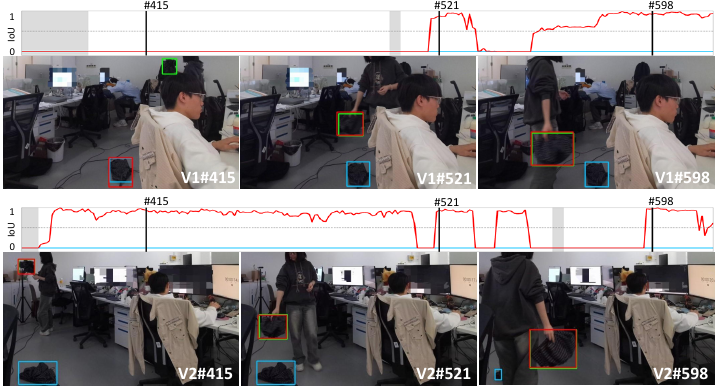}
		\caption{Two views: \textit{umbrella2-1} and \textit{umbrella2-2}. Under the interference of a similar object, ODTrack fails to re-track the correct target. In contrast, with the aid of multi-view assistance, MITracker can correct tracking errors from frame V1\#415 to \#521.}
		\label{fig: umbrella2}
        \end{subfigure}

        \vspace{3mm}
        \centering
        \begin{subfigure}{0.8\textwidth}
		\centering
		\includegraphics[width=\textwidth]{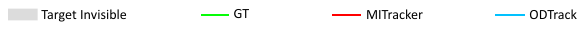}
        \end{subfigure}
        
	\caption{Qualitative comparison results on the MVTrack dataset using ODTrack.}
	\label{fig: pingpong5 and umbrella2}
\end{figure*}

\begin{figure*}[th]
	\centering
	\begin{subfigure}{0.9\textwidth}
		\centering
		\includegraphics[width=\textwidth]{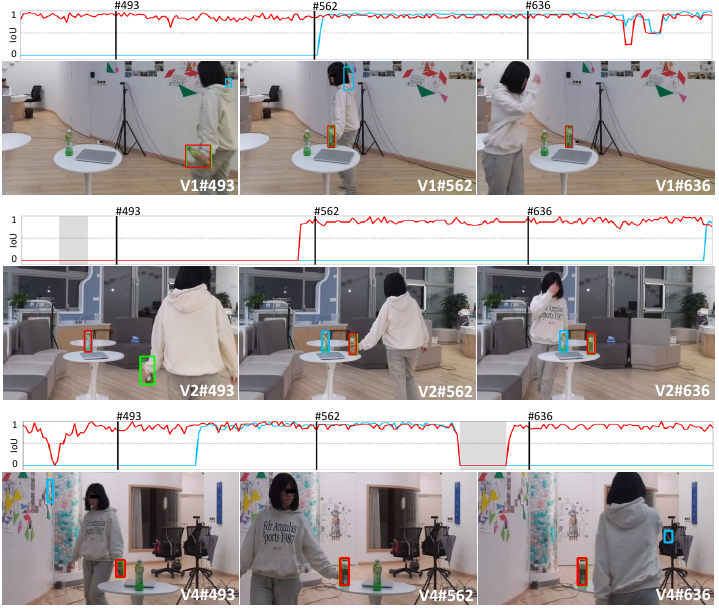}
	\end{subfigure}

        \centering
        \begin{subfigure}{0.9\textwidth}
		\centering
		\includegraphics[width=0.89\textwidth]{images/appendix/method_odtrack.pdf}
        \caption{Three views: \textit{bottle3-1},  \textit{bottle3-2} and \textit{bottle3-4}. In V2 \#493, MITracker momentarily mistracks a similar object as the target but successfully re-tracks the target by \#562. In contrast, ODTrack struggles to recover once it mistracks.}
		\label{fig: bottle3}
        \end{subfigure}

        \vspace{3mm}
        \begin{subfigure}{0.9\textwidth}
		\centering
		\includegraphics[width=\textwidth]{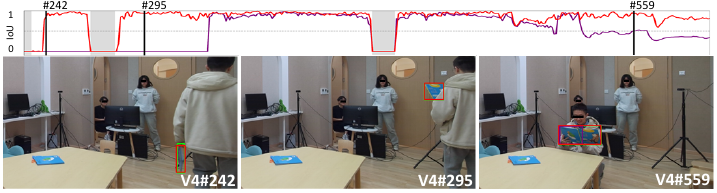}
        \end{subfigure}
        
        \centering
        \begin{subfigure}{0.9\textwidth}
		\centering
		\includegraphics[width=0.89\textwidth]{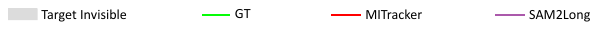}
        \caption{Sequence: \textit{book4-4}. SAM2Long completely loses the target following disappearances at frames \#242 and \#295. Upon re-tracking, it fails to adapt to target deformation, resulting in diminished IoU by frame \#559.}
		\label{fig: book4}
        \end{subfigure}

	\caption{Qualitative comparison results on the MVTrack dataset using ODTrack and SAM2Long.}
	\label{fig: bottle3 and book4}
\end{figure*}

\begin{figure*}[th]
	\centering
	\begin{subfigure}{0.9\textwidth}
		\centering
		\includegraphics[width=\textwidth]{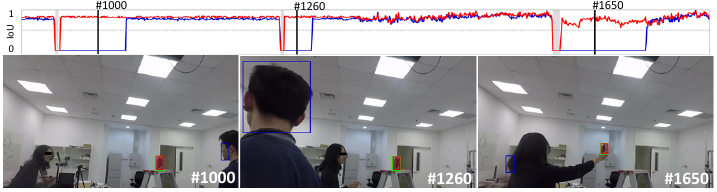}
		\caption{Sequence: \textit{cola-2}. MITracker demonstrates faster re-tracking capabilities than EVPTrack upon target reappearance.}
		\label{fig: cola-2}
	\end{subfigure}

        \vspace{3mm}
        \begin{subfigure}{0.9\textwidth}
		\centering
		\includegraphics[width=\textwidth]{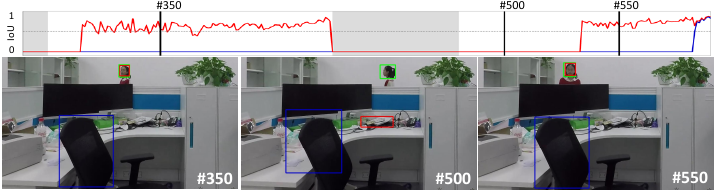}
		\caption{Sequence: \textit{manInOffice-2}. EVPTrack fails to correct after mistracking. In contrast, MITracker exhibits superior recovery capabilities, as demonstrated between frames \#500 and \#550.}
		\label{fig: manInOffice-2}
        \end{subfigure}

        \vspace{3mm}
        \centering
        \begin{subfigure}{0.8\textwidth}
		\centering
		\includegraphics[width=\textwidth]{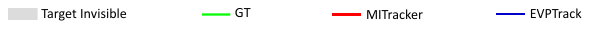}
        \end{subfigure}
        
	\caption{Qualitative comparison results on the GMTD using EVPTrack.}
	\label{fig: QGMTD}
\end{figure*}

We provide additional visual comparison results as illustrated in Figure \ref{fig: pingpong5 and umbrella2} and Figure \ref{fig: bottle3 and book4} from the MVTrack dataset, and Figure \ref{fig: QGMTD} from the GMTD. MITracker exhibits enhanced re-tracking capabilities both in multi-view and single-view scenarios. Furthermore, multi-view information assists in correcting instances of mistracking. 
To facilitate better visualization, each frame is cropped to a fixed area. The IoU curves above further illustrate the tracking accuracy by comparing each method’s predictions to the ground truth.

%% file: tables/dataset/attributes_table.tex
\begin{table}[h]
% \scriptsize
\begin{tabularx}{\linewidth}{l | X}
\bottomrule
\textbf{Att.} &\textbf{Definition}\\
\hline
\textbf{BC} & The background has similar appearance as the target\\
\textbf{MB} & The target region is blurred due to target motion\\
\textbf{POC} & The target is partially occluded in the frame\\
\textbf{FOC} & The target is fully occluded in the frame\\
\textbf{OV} & The target completely leaves the video frame\\
\textbf{DEF} & The target is deformable during tracking\\
\textbf{LR} & The target BBox is smaller than 1000 pixels\\
\textbf{ARC} & The ratio of BBox aspect ratio is outside the range [0.5, 2]\\
\textbf{SV} & The ratio of BBox is outside the range [0.5, 2]\\
\bottomrule
\end{tabularx}
\caption{Description of 9 attributes in MVTrack dataset.}
\label{tab: attributes description}
\end{table}